\documentclass{article}
\usepackage{amssymb}

\usepackage[final]{corl_2025} 

\usepackage{xcolor}
\newcommand{\paraDraft}[1]{\ifdefined\draft\subsubsection*{\color{blue}\textbf{#1}}\fi}

\usepackage{amsmath}
\usepackage{amsfonts}
\usepackage{mathtools}
\usepackage{siunitx} 

\providecommand{\eg}{\textit{e.g.,}~} %
\providecommand{\ie}{\textit{i.e.,}~} %
\usepackage{graphicx}   
\usepackage{caption}
\usepackage{subcaption}  
\usepackage{bm} 
\usepackage{multirow}
\usepackage{makecell} 

\usepackage{bm}
\usepackage[capitalise]{cleveref}
\crefname{appendix}{Appx.}{Appxs.}
\definecolor{infoRed}{RGB}{239,41,61} 
\definecolor{infoBlue}{RGB}{12,46,89} 
\colorlet{mygreen}{black!50!green}
\colorlet{myyellow}{black!25!yellow}
\definecolor{marinaRed}{RGB}{228,26,28} 
\definecolor{marinaBlue}{RGB}{55, 126, 184} 
\definecolor{marinaGreen}{RGB}{77, 175, 74} 
\definecolor{marinaPurple}{RGB}{152, 78, 163} 
\definecolor{marinaOrange}{RGB}{255, 127, 0} 
\definecolor{marinaYellow}{RGB}{255, 255, 51} 
\definecolor{marinaBrown}{RGB}{166, 86, 40} 
\definecolor{marinaPink}{RGB}{247, 129, 191} 
\ifdefined\draftTikz
    \usepackage{tikz}%
    \usepackage{graphicx}
    \usetikzlibrary{external}
    \usetikzlibrary{calc, fadings, patterns, arrows, fit, backgrounds, intersections, arrows.meta, bending, positioning, shapes.geometric}%
    \tikzset{%
        font=\footnotesize,
        frame/.style={%
            rectangle, draw,
            text width=6em, text centered,
            minimum height=3.6em,
            fill=white,
            rounded corners,
        },
        line/.style={%
            draw, -{Latex},
        }
    }
    \usepackage{pgfplots}%
    \usepackage{pgfplotstable}%
    \usepgfplotslibrary{groupplots, fillbetween}%
    \pgfplotsset{%
        compat=newest,
        grid=both,
        tick align=outside,
        tick pos=left,
        major tick length=2pt,
        every tick/.style={%
            black,
            thin,
        },
        enlargelimits=false,
        every axis plot/.append style={%
            ultra thick,
            font=\scriptsize,
        }, 
        legend cell align={left},
        every axis legend/.append style={%
            font=\scriptsize,
            rounded corners=2.0pt,
            draw=gray,
            fill opacity=0.8,
        },
        every tick label/.append style={%
            font=\scriptsize,
            /pgf/number format/.cd,
            fixed relative,
            precision=2,
            zerofill,
            /tikz/.cd,
        },
        label style={font=\scriptsize},
    }
\fi

\usepackage{hyperref}

\hypersetup{
    colorlinks=true,
    citecolor=infoRed,   
    linkcolor=infoRed,
    urlcolor=infoRed
}

\usepackage{wrapfig} 
\usepackage{pifont} 
\newcommand{\cmark}{\textcolor{green}{\ding{51}}} 
\newcommand{\xmark}{\textcolor{red}{\ding{55}}} 
\usepackage{appendix}  

\usepackage[capitalise]{cleveref}

\title{Poke and Strike: Learning Task-Informed Exploration Policies}

\author{
  Marina Y.~Aoyama$^{1}$, Jo\~{a}o Moura$^{1}$, Juan Del Aguila Ferrandis$^{1}$, Sethu Vijayakumar$^{1}$\\
  $^{1}$School of Informatics, The University of Edinburgh, UK\\
}

\begin{document}
\maketitle


\begin{abstract}
    In many dynamic robotic tasks, such as striking pucks into a goal outside the reachable workspace, the robot must first identify the relevant physical properties of the object for successful task execution, as it is unable to recover from failure or retry without human intervention.
    To address this challenge, we propose a task-informed exploration approach, based on reinforcement learning, that trains an exploration policy using rewards automatically generated from the sensitivity of a privileged task policy to errors in estimated properties. 
    We also introduce an uncertainty-based mechanism to determine when to transition from exploration to task execution, ensuring sufficient property estimation accuracy with minimal exploration time. 
    Our method achieves a 90\% success rate on the striking task with an average exploration time under 1.2 seconds---significantly outperforming baselines that achieve at most 40\% success or require inefficient querying and retraining in a simulator at test time. 
    Additionally, we demonstrate that our task-informed exploration rewards capture the relative importance of physical properties in two manipulation tasks and the classical CartPole example. 
    Finally, we validate our approach by demonstrating its ability to identify object properties and adjust task execution in a physical setup using the KUKA iiwa robot arm. 
    The project website is available at~\href{https://marina-aoyama.github.io/poke-and-strike/}{\textcolor{red}marina-aoyama.github.io/poke-and-strike/}. 
\end{abstract}



\section{Introduction}
    \paraDraft{Problem and motivation}
    Exploratory motions are essential for identifying system parameters and adjusting actions accordingly when executing interactive tasks. 
    Unlike visual properties, identifying physical properties---such as friction, mass, weight distribution, and restitution---requires active interaction with objects. 
    For example, a robot might push an object to identify friction for striking~\citep{memmel2024asid}, poke a sponge to evaluate stiffness for wiping~\citep{aoyama2024few}, or shake~\citep{saito2019real} or stir~\citep{guevara2020stir} liquids to determine viscosity for pouring.
    While such exploratory behaviors naturally emerge in humans~\citep{lederman1987hand}, enabling robots to autonomously explore through physical interactions remains a significant challenge.

    \textbf{Task-informed exploration.} 
    A key challenge in achieving exploratory behaviors is coming up with informative exploration strategies that identify task-relevant properties. 
    Most existing approaches rely on pre-defined, human-designed exploratory motions tailored to specific tasks~\cite{saito2019real, guevara2020stir, tai2023scone, aoyama2024few, antonova2022abayesian, saito2021how}, which can be suboptimal and cumbersome to obtain for every new task.
    Alternatively,~\citet{memmel2024asid} propose learning informative exploratory motions given a set of properties of interest, while~\citet{liang2020learning} introduce exploration policy learning guided by task information using reinforcement learning (RL), implicitly prioritizing the estimation of task-relevant properties. 
    However, both approaches~\cite{memmel2024asid, liang2020learning} require inefficient simulator queries and re-optimization of task motions for each object with the identified properties at test time. 
    Our approach, in contrast, enables immediate task execution after exploration, in addition to learning exploratory motions to identify task-relevant properties using rewards automatically generated from the task policy. 
    

    \textbf{Uncertainty-based policy switching.} 
    Another challenge lies in the transition from exploration to task execution. 
    Prior approaches often rely on a fixed-length exploration phase~\cite{memmel2024asid, liang2020learning}, which may be either too short, leaving the robot with insufficient information, or excessively long, which is inefficient. 
    To autonomously switch from exploration to task execution, the robot needs sufficiently accurate property estimates for a given task.
    However, the estimation accuracy is inaccessible outside of the simulated environment.
    Additionally, the robot must determine how accurately it needs to estimate each property to ensure task success. 
    To address this, our approach computes uncertainty estimates and, in simulation, we determine the uncertainty thresholds based on task completion. 

    In summary, the contributions of this work are:
    \begin{itemize} 
        \item We propose a task-informed exploration RL approach that trains an exploration policy using rewards automatically generated from the sensitivity of a privileged task policy to errors in estimated properties, leading to the identification of task-relevant properties. 
        \item We demonstrate that autonomously discovered exploration rewards lead to improved task performance by more accurately identifying relevant properties.
        \item We introduce the simultaneous learning of an online estimator with the exploration policy, enabling estimation within distribution of the exploratory observed states.
        \item  We present a method that autonomously transitions from exploration to task execution using uncertainty in property estimates, with thresholds computed from task success. 
        \item We validate our approach on a physical robotic setup, showing that it successfully adjusts task motions for different object properties. 
    \end{itemize}



    \begin{wrapfigure}{r}{0.5\textwidth}
        \vspace{-0.45cm}
        \centering
        \includegraphics[width=\linewidth]{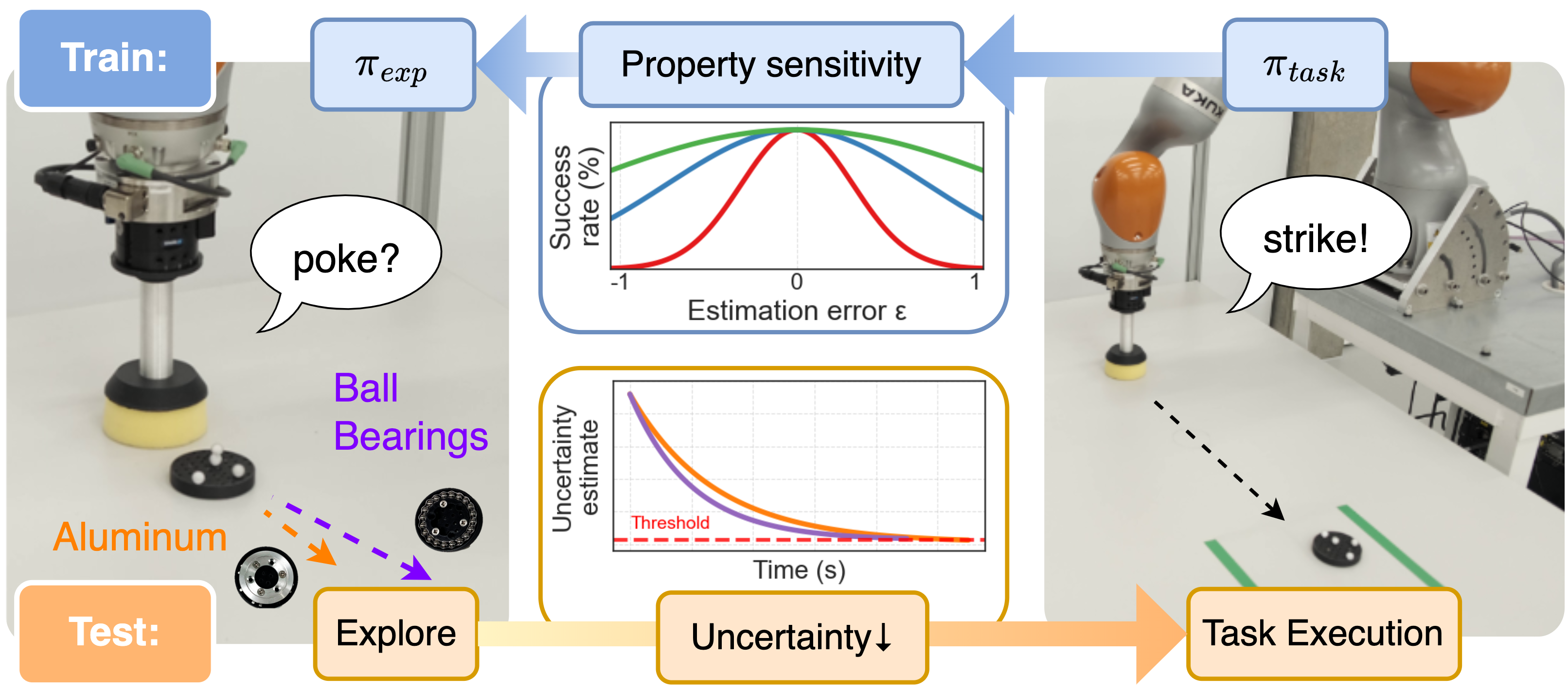}
        \caption{%
            \textbf{Task-informed exploration} approach enables the robot to autonomously learn how to explore and identify task-relevant properties by modeling task sensitivity to each property. 
            For dynamic tasks such as striking, the robot must first identify the object's properties through exploratory motions to achieve the task success, avoiding irreversible failure. 
        }
        \label{f:robot_sliding}
    \end{wrapfigure}

\section{Problem Statement}
    We address the problem of performing \textbf{one-shot robotic tasks} that involve interacting with objects or environments with \textbf{unknown physical properties}, where the robot is \textbf{unable to recover from failure or retry} without human intervention. 
    Examples include striking pucks with varying friction to an unreachable goal, tossing objects with different centers of mass, or shooting a basketball with unknown restitution. 
    These tasks require \textbf{identifying the physical properties of the manipulated object prior to task execution}, as initiating the task with incorrect assumptions about the properties can lead to irreversible failure---making online adaptation or correction unsuitable. 

    In this work, we seek to develop a method that addresses the following questions:
    (1) how to automatically discover informative exploratory motions for identifying task-relevant properties, when these motions differ from optimal task motion and vary depending on the task; 
    (2) how to adaptively determine the timing to transition from exploration to task execution, ensuring sufficient property estimation accuracy for a given task while minimizing exploration time; and 
    (3) how to execute the task immediately after exploration, without additional simulation queries or retraining at test time. 

\section{Related Work}
    \textbf{Domain randomization.}
    Domain randomization~\cite{tobin2017domain, akkaya2019solving} is a powerful technique in RL for enhancing the robustness of machine learning models and policies by introducing variability in environmental factors, such as the visual and physical properties of objects. 
    While this approach improves the robustness of learnt policies across diverse environmental conditions, it often results in policies that exhibit average behavior~\cite{ramos2019bayessim, memmel2024asid}. 
    To adapt behavior based on the physical properties of objects, \citet{ramos2019bayessim} propose non-uniform domain randomization. 
    However, this method requires obtaining the posterior distribution of object properties and retraining the task policy for those specific properties each time the robot executes the task. 
    On the other hand, \citet{peng2018sim} use Recurrent Deterministic Policy Gradient (RDPG) in RL, aiming to infer dynamics by capturing temporal dependencies. 
    While inspired by these methods, we find that they alone are insufficient for tasks requiring exploratory motions to identify physical properties prior to task execution. 
    
    \textbf{System Identification.} 
    System identification~\cite{ljung1999system} is crucial for modeling system dynamics from observed data. 
    A key challenge is generating exploratory motions that yield informative observations for parameter identification. 
    Many prior approaches rely on predefined motions~\cite{lim2022real, antonova2022abayesian, tai2023scone, saito2021how, saito2019real, guevara2020stir, kolev2015physically}. 
    Others iteratively update parameters through repeated task executions~\cite{ren2023adaptsim, cheng2024iterative}.     
    However, optimal task motions may differ from those most informative for parameter identification. 
    To address this, prior works optimize exploratory motions by minimizing property estimation errors~\cite{kumar2019estimating}, reducing state transition prediction errors~\cite{zhou2018environment}, or maximizing Fisher information to generate observations that most effectively distinguish differences in properties~\cite{memmel2024asid}. 
    However, these methods assume equal importance of all given parameters and fail to identify task-relevant ones. 
    
    \citet{liang2020learning} introduce task-oriented exploration using RL, where task rewards implicitly guide the exploration policy to focus on properties critical for task success.
    However, their approach incurs inefficient simulation queries and requires a fixed exploration duration to infer parameters by matching simulated and real-world observations, aiming to mitigate the out-of-distribution issue with learned estimators~\cite{kumar2019estimating}. 
    In contrast, our method trains the online property estimator simultaneously with the exploration policy to align the training data distribution, while enabling adaptive exploration length based on uncertainty estimates. 
    Additionally, these prior works~\cite{liang2020learning, memmel2024asid} require re-optimization of the task policy at test time. Our method enables immediate task execution after exploration using an online property estimator and a privileged task policy.
    \citet{dass2024learning} also propose a dual-policy framework for task and information-seeking, but assume minimal action space overlap---precluding applicability in scenarios where exploratory and task actions interfere, such as physical property identification.
    In contrast, as summarized in~\cref{t:relatedwork_comparison}, our method enables one-shot task execution immediately after exploration, with automatically determined exploration time, and without constraints on action spaces.
    \begin{wraptable}{r}{0.54\textwidth}
        \centering
        \scriptsize
        \setlength{\arrayrulewidth}{0.2mm}
        \setlength{\tabcolsep}{2pt}
        \renewcommand{\arraystretch}{1.15}
        \begin{tabular}{
            >{\centering\arraybackslash}m{1.9cm} 
            >{\centering\arraybackslash}m{1.2cm} 
            >{\centering\arraybackslash}m{1.2cm} 
            >{\centering\arraybackslash}m{1.2cm} 
            >{\centering\arraybackslash}m{1.2cm}}
            \noalign{\hrule height 0.3mm}
            \vspace{2pt}Method\vspace{2pt} & 
            \vspace{3pt}\shortstack{One-shot\\task}\vspace{1pt} & \vspace{3pt}\shortstack{Immediate\\task\\execution}\vspace{1pt} & 
            \vspace{3pt}\shortstack{Adaptive\\exploration\\time}\vspace{1pt} &
            \vspace{3pt}\shortstack{Overlapping\\action\\space}\vspace{1pt} \\
            \hline
            BayesSim~\cite{ramos2019bayessim}       & \xmark & \xmark & \xmark & \cmark \\
            AdaptSim~\cite{ren2023adaptsim}         & \xmark & \cmark & \xmark & \cmark \\
            IRP~\cite{cheng2024iterative}         & \xmark & \cmark & \xmark & \cmark \\
            DISaM~\cite{dass2024learning}           & \xmark & \cmark & \cmark & \xmark \\
            Task-Oriented~\cite{liang2020learning}  & \cmark & \xmark & \xmark & \cmark \\
            ASID~\cite{memmel2024asid}              & \cmark & \xmark & \xmark & \cmark \\
            \textbf{Ours}                           & \cmark & \cmark & \cmark & \cmark \\
            \noalign{\hrule height 0.3mm}
        \end{tabular}
        \caption{
            \textbf{One-shot task} must succeed in a single attempt.  
            \textbf{Immediate task execution} after exploration, without simulator queries or policy retraining. 
            \textbf{Adaptive exploration time} automatically adjusted for sufficient but minimal exploration.  
            \textbf{Overlapping action space} shared between exploration and task execution. 
        }
        \label{t:relatedwork_comparison}
        \vspace{-2em}
    \end{wraptable}

    \textbf{Privileged learning for robot control.}
    Privileged learning~\cite{pechyony2010on} leverages information inaccessible during real-world deployment. 
    This privileged information accelerates the learning process by providing the agent with critical knowledge, such as hidden states or ground truth parameters, which would otherwise require extensive exploration to uncover. 
    \citet{chen2020learning} and subsequent works~\cite{fuchioka2024robotic, Lee2020learning, qi2023general, lin2022learning} proposed training a student policy to imitate the behavior of a privileged teacher policy using partial observations.  
    \citet{ferrandis2024learning} use a privileged policy to generate data for learning a state estimator, and \citet{yu2017preparing} use for a property estimator; 
    others incorporate privileged information into auxiliary losses~\cite{hu2024privileged, rostel2023estimator} or the critic network~\cite{pinto2018asymmetric, hu2024privileged}. 
    While most existing work in these settings focuses on identifying hidden states~\cite{ferrandis2024learning, fuchioka2024robotic, Lee2020learning, qi2023general, lin2022learning, rostel2023estimator} or parameters~\cite{Lee2020learning, qi2023general, yu2017preparing} from passive observation during task execution, our work is the first to address the challenge of uncovering parameters, specifically physical property parameters in our case, that require active exploration motions different from task motions. 
                            

\section{Method}
    We propose a task-informed exploration approach. 
    The core idea is to leverage the privileged task policy to automatically generate rewards based on its sensitivity to errors in estimated properties. 
    These rewards guide the training of the exploration policy to identify task-relevant properties, while simultaneously training the online property and uncertainty estimator.
    At deployment, the robot begins with the exploration policy, uses uncertainty estimates to transition to task execution, and immediately executes the task policy after exploration, adjusting its motion based on the final property estimates from the exploration. 
    ~\cref{subsec:appendix_framework} outlines a detailed overview of the proposed method.

    \subsection{Problem Formulation}
    We formulate the problem as a family of Markov Decision Processes (MDPs) where the dynamics of the environment depends on the physical properties $\phi \in \Phi$ of the manipulated object. 
    We assume that $\phi$ remains constant throughout the episode. 
    Each instance takes the form of the tuple $M(\phi) = (S, A, P_{\phi}, R, \gamma)$, with a finite horizon $H$, where $S$ is the state space, $A$ is the action space, $P_{\phi}$ is the transition model conditioned on $\phi$, $R$ is the reward function and $\gamma \in [0, 1)$ is the discount factor. 
    The objective is to learn a policy $\pi$ that maximizes the expected discounted return across different values of the physical properties $\phi$: 
    $\max_{\substack{\pi}} \mathbb{E}_{\phi \sim p(\phi), \pi, P_{\phi}} \left[ \sum_{t=0}^{H-1} \gamma^t R(s_t, a_t, s_{t+1}) \right]$. 
    The physical properties $\phi$ of the object are hidden, and the robot must infer them for successful task execution. 
    Since the one-shot nature of the task precludes online adaptation or retrying the task, the robot must first explore the object to infer these properties and then adjust its task execution accordingly.

    \subsection{Privileged Task Policy Learning}\label{s:privileged_policy_method} 
    \paraDraft{Motivation}
    We first train a privileged task policy $\pi_{\text{task}}$ using RL with access to ground truth physical property parameters $\phi^*$ in simulation. 
    We hypothesize that leveraging this privileged information reduces the need for extensive exploration and prevents convergence to suboptimal policies, leading to a higher success rate by adapting to different properties.
    
    \textbf{Task policy.}
    During training, the task policy receives the ground truth physical property values $\phi^*$ as input, along with the current state observation $s_t$ and the goal $g$ in goal-conditioned settings, and computes actions $a_t$ at each timestep. 
    We define the task policy reward, $r_{\text{task}}$, based on the task's objectives, with positive rewards for achieving desired behaviors, such as reaching a target, and negative rewards for undesirable ones, such as violating constraints or workspace boundaries. 
    We randomize the initial state to enable safe task execution in scenes perturbed by exploration. 

    \subsection{Learning of Exploration and Property Estimation}
    \paraDraft{Motivation and Summary}
    While privileged access to ground truth physical property values aids task policy training, these values are unavailable in real-world settings. 
    To address this, we simultaneously train (1) an exploration policy $\pi_{exp}$ to perform motions that are informative for estimating the physical properties $\phi$ prior to task policy execution, and (2) an online property estimator $f_\phi$ to infer these properties from state observations during exploration. 
    This estimator allows immediate task execution after exploration---unlike prior works~\cite{liang2020learning,memmel2024asid} that require querying a simulator for offline inference. 
    
    \textbf{Exploration policy.} 
    Next, we train an exploration policy $\pi_{\text{exp}}$ that computes actions $a_t$ to obtain observations informative for identifying the physical properties of the manipulated object, given the current states $s_t$ of the robot and the object through RL. 
    We define the exploration policy reward as 
    \begin{equation}
    r_{\text{exp}} = 
    \begin{cases}
    r_{\text{estimation}} & \text{if } \forall j \, \varepsilon_{\text{estimation},j} < \varepsilon_{\text{threshold},j} \\
    r_{\text{failure}} & \text{otherwise}
    \end{cases},
    \label{eq:exploration_reward_function}
    \end{equation}
    \noindent where $\varepsilon_{\text{estimation},j}$ represents the estimation error for the $j$-th physical property, and $\varepsilon_{\text{threshold},j}$ is the threshold for the $j$-th property. 
    The robot receives a positive reward $r_{\text{estimation}}$ if the estimation errors of all properties are below their respective thresholds. 
    We obtain the estimation errors $\varepsilon_\text{estimation}$ by computing the difference between the ground truth physical property values $\phi^*$ and the estimated physical property values $\hat{\phi}$ estimated by the physical property estimator~$f_\phi$, as
    \begin{equation}
        \varepsilon_{\text{estimation},j} = \left| \phi_j^\ast - \hat{\phi_j} \right|.
    \end{equation}
    We adopt on-policy RL for stable training under non-stationary rewards caused by dependence on the simultaneously trained estimator. 
    The robot receives a negative reward $r_{\text{failure}}$ for violating the workspace boundaries, ensuring feasible task execution after exploration, without manual reset. 
    
    \textbf{Online physical property estimator.}
    For online physical property estimation during exploration, we employ a Long Short-Term Memory (LSTM)~\cite{hochreiter1997long}, as temporal information is essential for capturing object dynamics (see~\cref{s:exploration_results} for alternative approaches, \eg a Transformer-based estimator). 
    Between each training update, the exploration policy collects a new dataset $\mathcal{D}$ of rollouts, consisting of states $s_t$ and the corresponding ground truth physical property values $\phi^\ast$. 
    At each update of the exploration policy, we also update the estimator by minimizing the estimation loss $\mathcal{L}_{est}$ (see~\cref{s:appendix_estimator_loss} for details) using the dataset $\mathcal{D}$.
    This simultaneous training of the estimator on observations from the most recent exploration policy mitigates out-of-distribution issues, a known challenge in learned estimators~\cite{kumar2019estimating} and forward models~\cite{memmel2024asid, shyam2019model}.

    \subsection{Task-Informed Exploration Reward Design}
    \label{s:exploration_rewards_method}
    \paraDraft{Motivation}
    
    The exploration policy reward function in ~\cref{eq:exploration_reward_function} requires specifying the estimation error thresholds~$\varepsilon_{\text{threshold},j}$. 
    While humans can leverage intuition and physics knowledge to identify which parameters are relevant to the target task, manually specifying optimal threshold values for multiple properties is non-intuitive. 
    Moreover, tuning these thresholds for each task is cumbersome, as the relevance of each physical property varies across tasks. 
    To address this, we propose a method to automatically generate these estimation thresholds by modeling task performance sensitivity to estimation error in each property. 
    Since we compute these thresholds from task performance sensitivity, these exploration rewards serve as a surrogate for the task reward. 
    This reward design enables the robot to learn exploratory motions that lead to high task performance, prioritizing the identification of task-relevant properties---without executing the task policy during exploration policy training. 
    
    \textbf{Modeling task performance sensitivity.} 
    For each physical property, we assume a uni-modal relationship between the task success rate $y$ and the estimation error $\varepsilon$, where task performance is highest when the estimation is accurate and decreases as the error increases. 
    The rate at which task success deteriorates with increasing error reflects the sensitivity of the task to each property. 
    
    To model this relationship, we fit a parametric uni-modal function $g_j(\varepsilon)$ to empirical data $\mathcal{D}_{\varepsilon,j}$ for each $j$-th physical property, where $g$ represents any uni-modal function. 
    The dataset $\mathcal{D}_{\varepsilon,j}$ consists of pairs of estimation error levels $\varepsilon$ and the corresponding task success rates $y$ achieved by the task policy. 
    During data collection, we roll out the privileged task policy, systematically replacing the ground truth value of the $j$-th physical property with perturbed values at varying levels of estimation error. 
    We model the sensitivity of each property individually by introducing errors into one property at a time while keeping all others unperturbed, assuming that the impact of each property on task performance is independent of the others.

    \textbf{Computing task-informed exploration rewards.}
    From the fitted uni-modal function $g_j$, we compute a set of estimation error thresholds $\varepsilon_{\text{threshold},j}$ for each physical property $j$, such that the task success rate remains above a proportion $p$ of the maximum success rate achieved by the privileged task policy. Specifically, we solve:
    \begin{equation}
    g_j(\varepsilon_{\text{threshold},j}) \geq p \cdot \max_{\varepsilon} g_j(\varepsilon)
    \label{eq:task-informed_exploration_reward}
    \end{equation}
    for each property. 
    These estimation error thresholds define the success criteria in the exploration reward in~\cref{eq:exploration_reward_function}. 
    Properties that are more relevant to the task result in tighter thresholds, encouraging the exploration policy to estimate those properties with higher accuracy. 
    
    \subsection{Uncertainty-Based Policy Switching}
    \paraDraft{Motivation}
    In the training phase, we define exploration success using estimation error, as in~\cref{eq:exploration_reward_function}.
    However, this error is inaccessible at test time on a physical setup. 
    To transition from exploration to task execution once the property estimates are sufficiently accurate, we 1) estimate the uncertainty of the property estimates, and 2) determine the uncertainty thresholds required for successful task execution by modeling the relationship between each property's uncertainty and the task outcome. 
    
    \textbf{Uncertainty estimator.}
    We estimate the predictive uncertainty of the physical property estimator $f_\phi$ using an ensemble approach that captures both aleatoric uncertainty (from data noise) and epistemic uncertainty (from model limitations) \cite{gawlikowski2023survey}.
    We define $f_\phi$ as an ensemble of~$M$ neural networks, indexed by~$i$, each comprising two heads: one for the predicted mean $\hat{\phi}_i(s_t)$ (denoted~$\hat{\phi}_{i,t}$) and another for the predicted covariance $\hat{\Sigma}_i(s_t)$ (denoted~$\hat{\Sigma}_{i,t}$) \cite{nix1994estimating, lakshminarayanan2017simple, russell2021multivariate}.
    Assuming a heteroscedastic setting, i.e.~the uncertainty depends on the models' input, where $p(\phi | s_t) = \mathcal{N}(\hat{\phi}_{i,t}, \hat{\Sigma}_{i,t})$, we train each model $i$ to minimize the negative log-likelihood.
    The ensemble consists of an equally weighted mixture of Gaussians which, for simplicity, we approximate with a single Gaussian distribution, such that $p(\phi | s_t) = \mathcal{N}(\hat{\phi}_t, \hat{\Sigma}_t)$, where the mean and covariance are those of the mixture \cite{lakshminarayanan2017simple}:
    \begin{equation*}
       \hat{\phi}_t = \frac{1}{M}\sum_{i=1}^M \hat{\phi}_{i,t}, \qquad
        \hat{\Sigma}_t =
        \underbrace{\frac{1}{M} \sum_{i=1}^M\hat{\Sigma}_{i,t}}_{\mathclap{\substack{\text{mean of the}\\\text{individual covariances}}}}
        +
        \underbrace{\frac{1}{M} \sum_{i=1}^M \left(\hat{\phi}_{i,t}\hat{\phi}^T_{i,t}\right) - \hat{\phi}_t\hat{\phi}^T_t}_{\mathclap{\substack{\text{covariance of the}\\\text{mixture means}}}}.
        \vspace{-0.1em}
    \end{equation*}
    We use the mean $\hat{\phi}_t$
    as the predicted physical property values, and the covariance $\hat{\Sigma}_t$
    as the measure for predictive uncertainty, where the mean of the individual covariances captures aleatoric uncertainty, while the covariance of the mixture means captures epistemic uncertainty.
    
    \textbf{Computing uncertainty thresholds.}
    Finally, we compute the uncertainty thresholds required for successful task execution. 
    We roll out the exploration and task policies to collect uncertainty data labeled with task outcomes (success or failure). 
    Then, we calculate the p-th percentile of the uncertainty values for each property from successful task trials (\ie $q\%$ of successful trials have uncertainty values lower than this threshold). 
    Therefore, the task policy is likely to succeed when the uncertainty values are below these thresholds. 
    These thresholds enable the robot to assess exploration success and switch to the task policy without direct access to estimation error during testing.     
    
\begin{wrapfigure}{r}{0.44\textwidth} 
        \centering    
        \vspace{-2em} 
        \begin{minipage}{0.22\textwidth}
            \centering
            \subcaption*{\texttt{Striking}}
            \includegraphics[width=\linewidth]{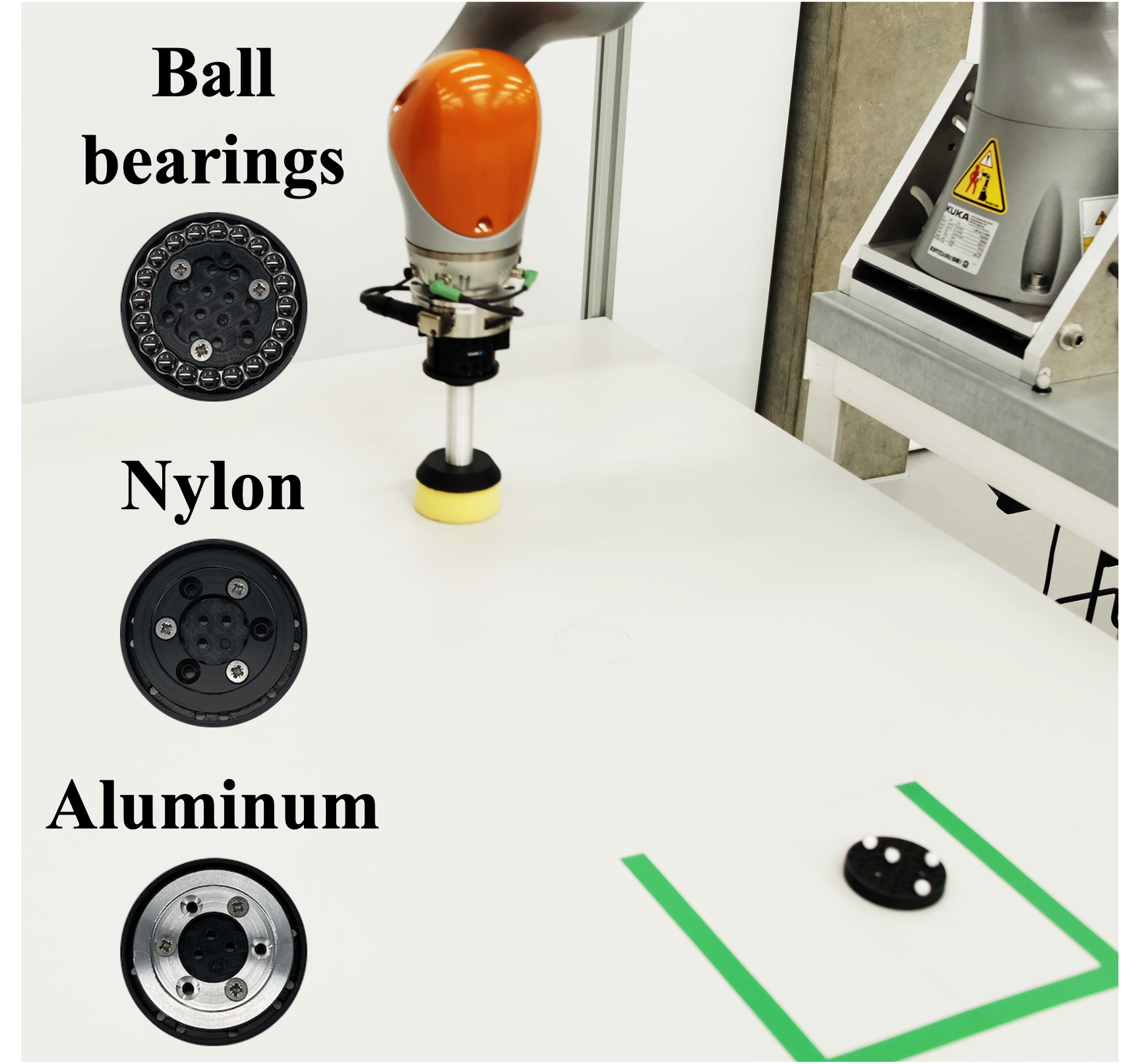}
        \end{minipage}%
        \hfill
        \begin{minipage}{0.22\textwidth}
            \centering
            \subcaption*{\texttt{Edge Pushing}}
            \includegraphics[width=\linewidth]{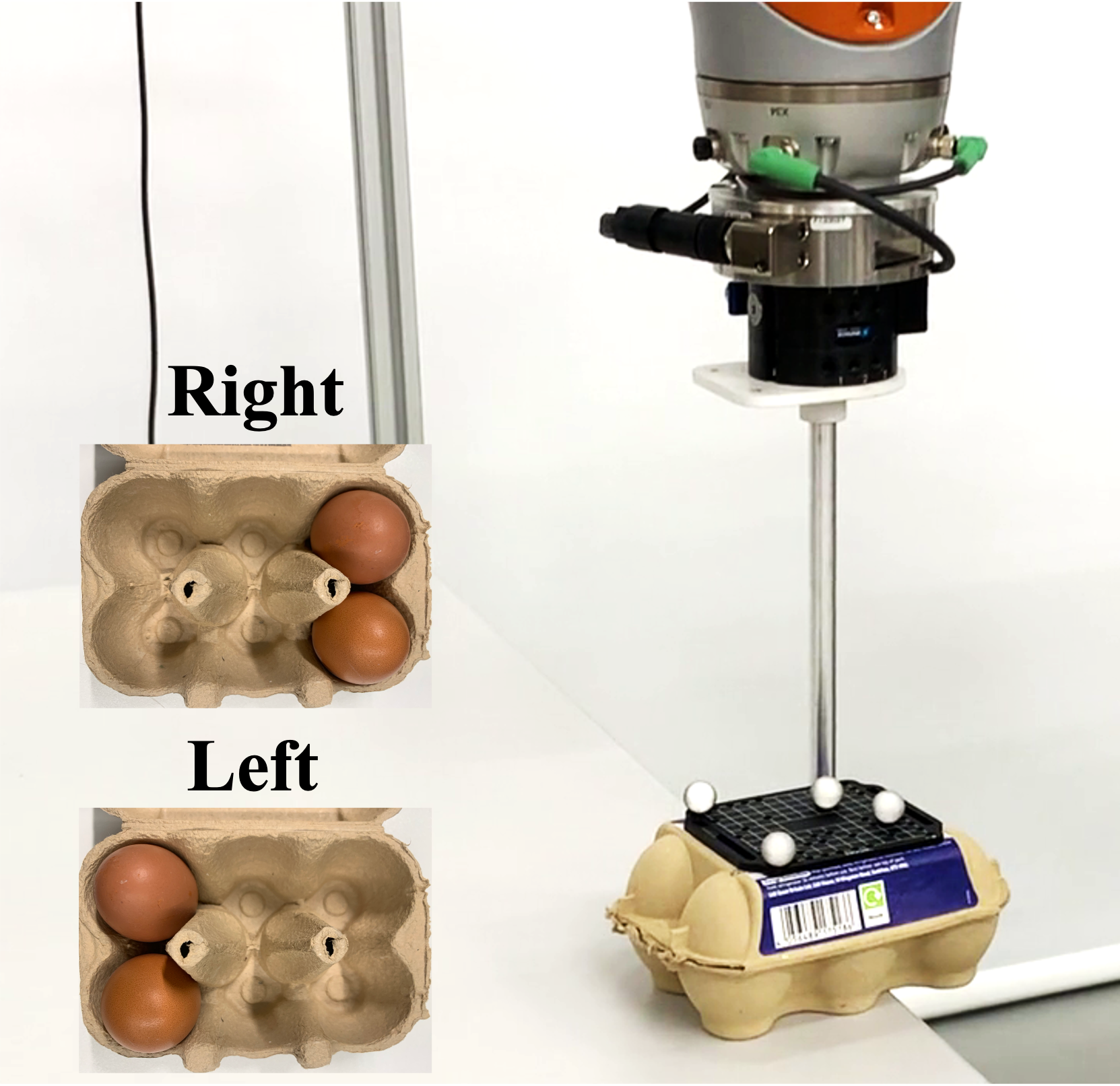}
        \end{minipage}
        \caption{Manipulation tasks. }
        \label{f:robot_tasks}
        \vspace{-2em} 
    \end{wrapfigure}
\section{Experiments}
    \paraDraft{Experiment overview} 
    We evaluate our task-informed exploration approach in simulation and on a physical robot. We provide additional implementation details in~\cref{s:appendix_experimental_details}. 

    \paraDraft{Task}
    \textbf{Tasks.} 
    We evaluate our method on three tasks: 
    \texttt{Striking} a puck with unknown physical properties toward an unreachable goal; 
    \texttt{Edge Pushing} a box with unknown contents to the table edge, where incorrect property estimates cause it to fall; and 
    the classic \texttt{CartPole} task. 
    Task details are in~\cref{subsec:tasks}. 
    
    \paraDraft{Baselines}
    \textbf{Baselines.} 
    We compare our method with the following baselines: 
    Domain Randomization (DR)~\cite{tobin2017domain}, which randomizes physical properties for generalization; 
    DR+Stack, which augments DR with stacked observations to capture the temporal dynamics of objects influenced by physical properties; 
    DR+LSTM, which integrates an LSTM into PPO to model temporal dependencies, similar to dynamic randomization with LSTM~\cite{peng2018sim} but using PPO instead of RDPG; 
    DR+Stack+Est, which extends DR+Stack with one head for action computation and another for property estimation; 
    Student~\cite{chen2020learning}, which trains a student policy using DAgger; 
    RMA~\cite{kumar2021rma}, which trains to encode observations by regressing towards the privileged latent representation of physical properties; and 
    UP-OSI~\cite{yu2017preparing}, which simultaneously trains a privileged policy and an online property estimator. 
     
     \subsection{Does the task-informed exploration approach improve task performance?} \label{s:main_comparison_results}
\begin{wrapfigure}{r}{0.45\textwidth} 
        \centering
        \includegraphics[width=0.45\textwidth, keepaspectratio]{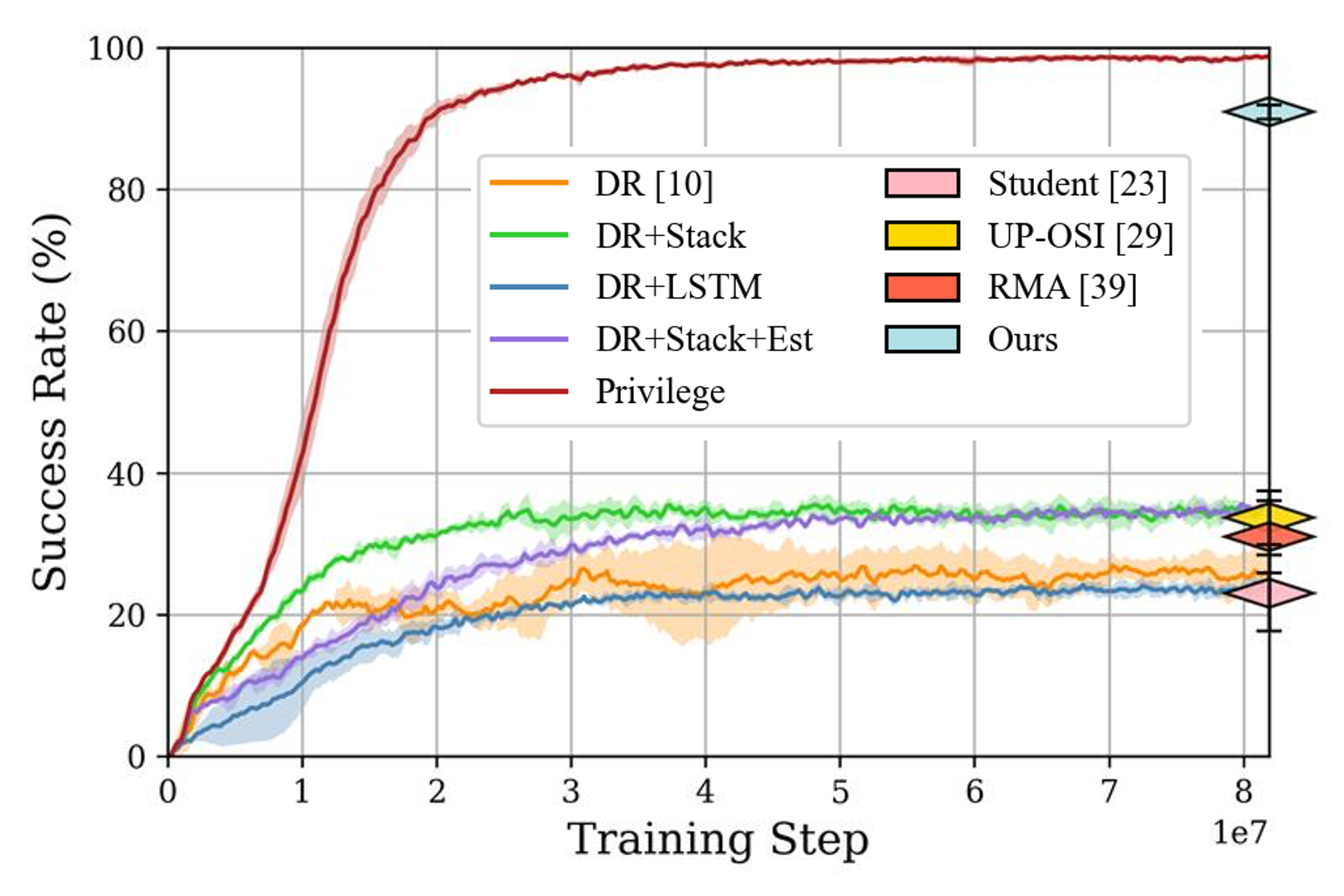}
        \caption{
            Performance of different methods on the \texttt{Striking} task, with mean and standard deviation reported across three training seeds.
        }
        \label{f:main_comparison}
        \vspace{-1.5em}
    \end{wrapfigure}

     \textbf{Ours.} 
     \cref{f:main_comparison} summarizes the results for the one-shot \texttt{Striking} task. 
     Our method achieves a 90.1\% success rate, significantly outperforming all baselines, which reach at most 40\%. 
     Our policies achieve 92.3\% success in exploration (see~\cref{s:exploration_results}) and 98.7\% in task, demonstrating that the exploration policy estimates properties with sufficient accuracy for successful task performance. 
     We confirmed consistent results on the \texttt{Edge Pushing} task (see~\cref{s:edgepushing_baseline_comparison}). 
    
     \textbf{DR baselines.} DR alone learns an average motion across properties and achieves only a 25.4\% success rate. 
     In methods with temporal information, DR+Stack and DR+LSTM, exploratory pushing motions emerge but attains only 35.4\% and 23.3\% task success, respectively. 
     These methods rely on delayed task rewards, provided only after task execution, to evaluate the effectiveness of the exploration, making it difficult to associate the exploratory motions with the rewards. 
     Adding rewards for property estimation, as in DR+Stack+Est, also achieves only 34.8\% due to the challenge of balancing exploration and task execution rewards within a single policy. 
     
     \textbf{Privileged baselines.} 
     The methods leveraging a privileged policy (Student, UP-OSI, and RMA) achieved success rates of 23.0\%, 33.7\%, and 31.0\%, respectively. 
    Since the privileged task policy lacks exploratory behavior, imitating it (Student) or rolling it out---whether using online estimation (UP-OSI) or latent encodings (RMA)---leads to task failure. 

     
    \begin{wrapfigure}{r}{0.5\textwidth} 
        \centering
        \vspace{-1em} 
        \includegraphics[width=0.45\textwidth]{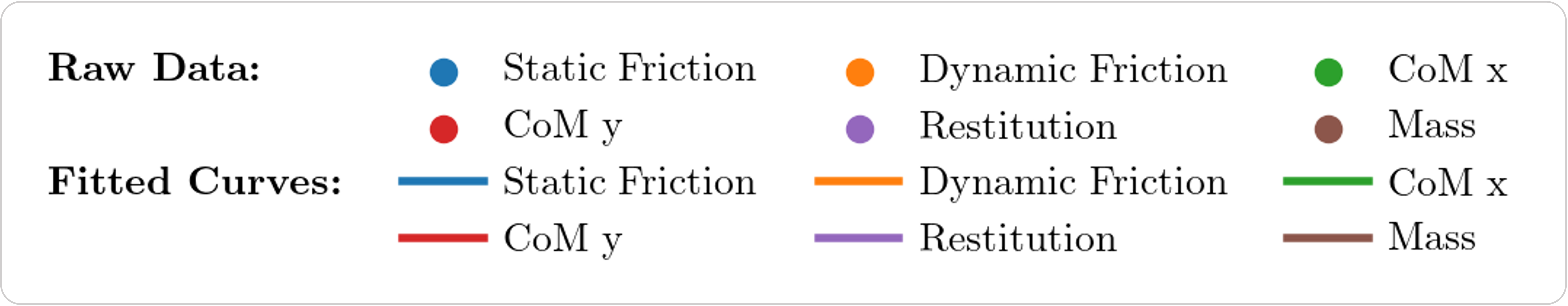}
        \vspace{-0.7em}
        \begin{minipage}{0.255\textwidth} 
            \subcaption*{\texttt{Striking}}
            \vspace{-0.2em}
            \includegraphics[width=\linewidth]{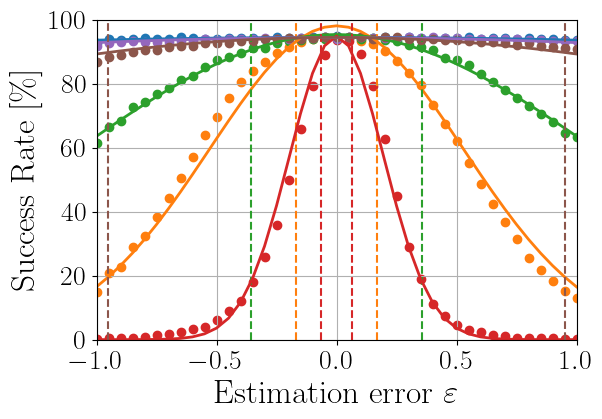}
            \label{fig:placeholder}
        \end{minipage}
        \begin{minipage}{0.23\textwidth} 
            \subcaption*{\texttt{Edge Pushing}}
            \vspace{-0.2em}
            \includegraphics[width=\linewidth]{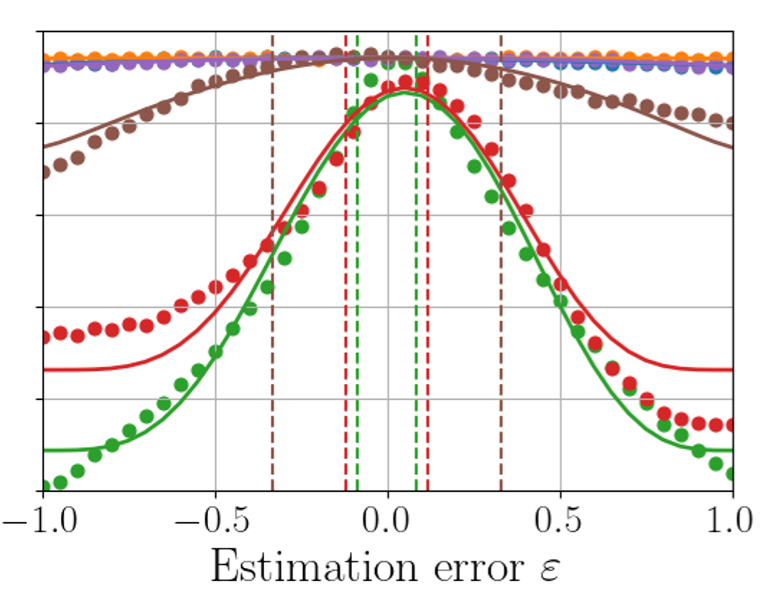}
            \label{fig:placeholder}
        \end{minipage}
        \caption{
            Uni-modal functions fitted to the relationship between task success rate and normalized property estimation errors, modeling task sensitivity. 
            Dashed lines indicate the computed estimation error thresholds. 
        }
        \label{f:uni-modal_main}
        \vspace{-2em}
    \end{wrapfigure}
    \subsection{Do task-informed exploration rewards capture task-relevant physical properties?}\label{s:sensitivity_results}
    \paraDraft{Modeling task performance sensitivity} 
    \cref{f:uni-modal_main} presents task sensitivity to errors in each physical property parameter. 
    In the \texttt{Striking} task, the CoM in the y-direction and dynamic friction show a sharper decline in success rate as error increases, compared to less sensitive parameters such as static friction.
    This sensitivity model indicates that the former parameters are more relevant for performing the task, resulting in tighter estimation thresholds for exploration rewards. 
    The results, aligning with our physics-based intuition, demonstrate that our method automatically identifies task-relevant properties.
    Similarly, in the \texttt{Edge Pushing} and \texttt{CartPole} tasks, each property exhibits distinct relationships, reflecting varying levels of relevance to task performance. 
    See~\cref{s:appendix_sensitivity} for further analysis on all tasks. 

    
    \subsection{How do task-informed exploration rewards and uncertainty estimates contribute?} 
    \paraDraft{Ablation explanation. }
    \textbf{Task-informed exploration rewards.}
    We compare the task-informed rewards from \cref{eq:task-informed_exploration_reward} against task-agnostic rewards on the \texttt{Striking} task. 
    When thresholds are uniformly too high (\ie $\epsilon = \max_j \epsilon_{\text{threshold}, j}$), task success drops to 47.3\%, failing to accurately estimate task-relevant properties. 
    When thresholds are uniformly too low (\ie $\epsilon = \min_j \epsilon_{\text{threshold}, j}$), exploration success drops to 8.0\%, requiring highly accurate identification of all properties. 
    In contrast, task-informed thresholds achieve 92.3\% exploration success and 90.1\% task success, supporting our hypothesis that computing the highest estimation thresholds (to trigger exploration rewards) while maintaining sufficiently low thresholds (to obtain more accurate estimates) is crucial. 
    See~\cref{s:appendix_ablation_table} for complete results.
    
    \begin{wrapfigure}{r}{0.39\textwidth} 
        \centering
        \vspace{-1.5em} 
        \begin{minipage}{0.24\textwidth} 
            \includegraphics[width=\linewidth]{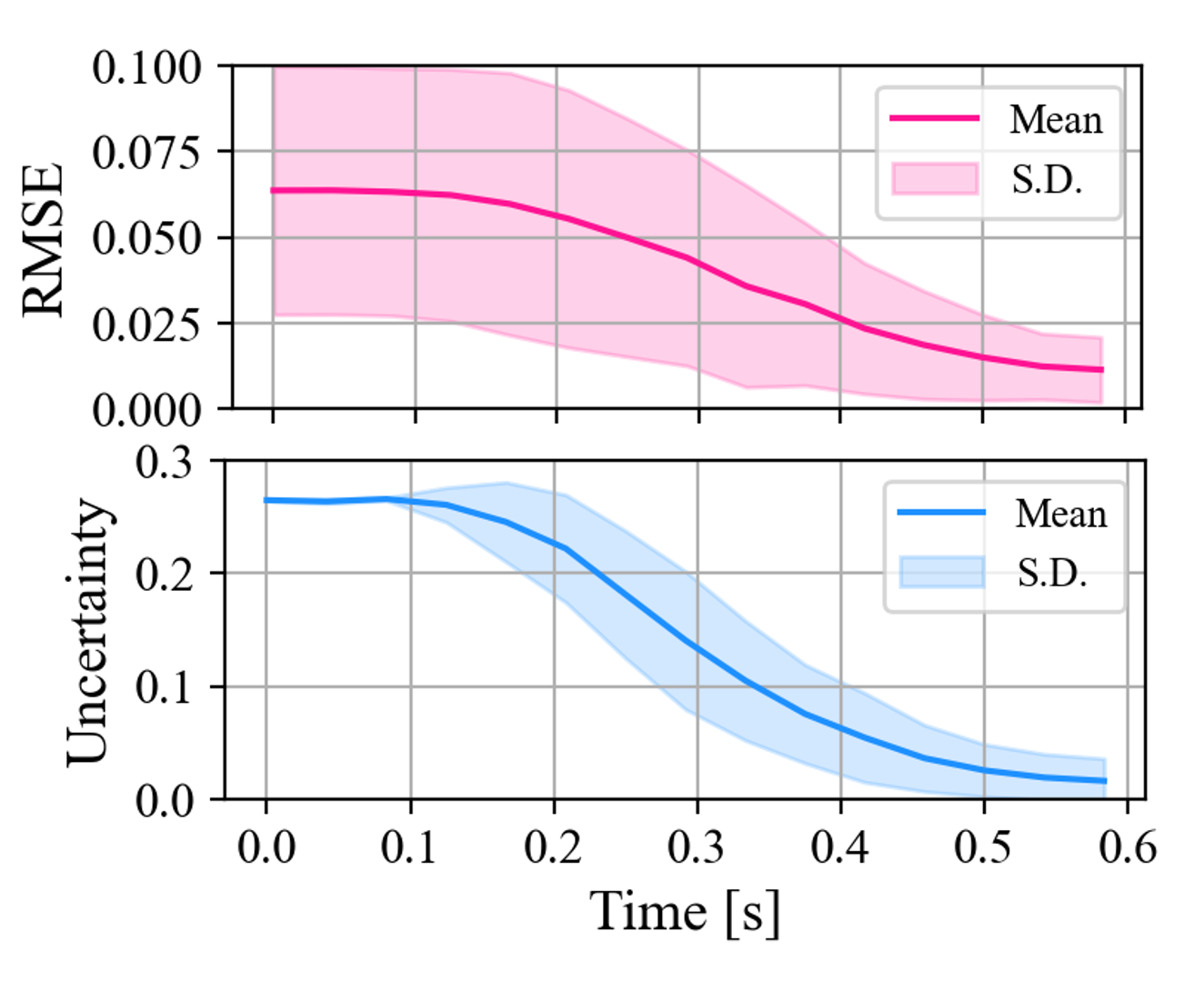}
            \vspace{-1.8em}
            \subcaption{}
            \label{fig:main_unc_est_time}
        \end{minipage}
        \hspace{-0.5em}
        \begin{minipage}{0.15\textwidth} 
            \includegraphics[width=\linewidth]{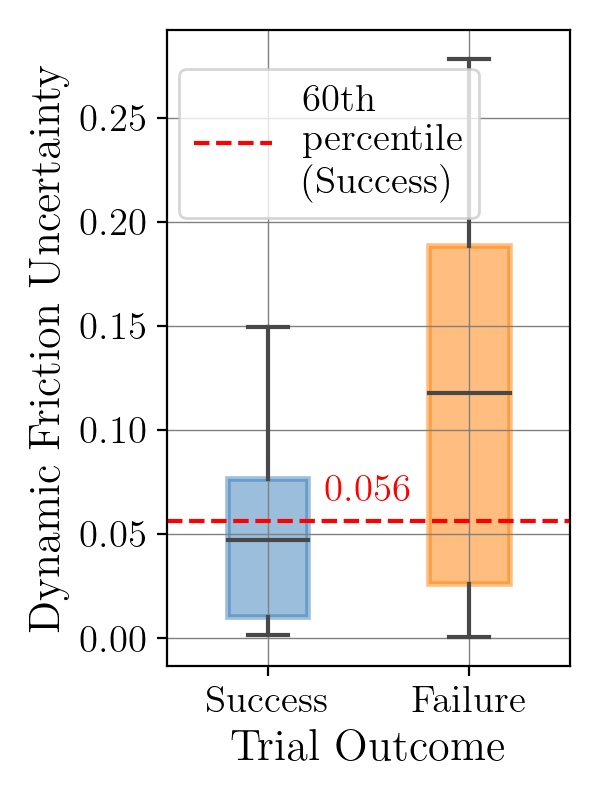}
            \vspace{-1.8em}
            \subcaption{}
            \label{fig:main_unc_success}
        \end{minipage}
        \vspace{-0.5em}
        \caption{
            (a) RMSE and uncertainty of dynamic friction over time. 
            (b) Uncertainty distribution for successful vs. failed trials.
        }
        \label{f:unc_analysis_main}
        \vspace{-2em}
    \end{wrapfigure}
    \textbf{Uncertainty estimates.} 
    \cref{fig:main_unc_est_time} shows that both estimation errors and uncertainties decrease as exploration progresses when rolling out the exploration policy 100 times for the \texttt{Striking} task. 
    Further,~\cref{fig:main_unc_success} shows that low uncertainty in task-relevant property estimates leads to task success in sim2sim transfer (\ie, rolling out in PyBullet). 
    The dotted red line indicates the computed uncertainty threshold for policy switching. 
    See~\cref{s:uncertainty_results} for a full analysis, including uncertainty estimates of task-irrelevant properties. 
    These findings support using uncertainty as a proxy for estimation error in physical setups and for determining when to switch to task execution.
    \begin{wrapfigure}{r}{0.45\columnwidth}
        \vspace{-2em}
        \centering
        \ifdefined\draftTikz
            \tikzsetnextfilename{fig_robot_exp_snapshot}
            \input{fig_robot_exp_snapshot.tex}
        \else
            \begin{subfigure}{\linewidth}
                \includegraphics[width=\linewidth]{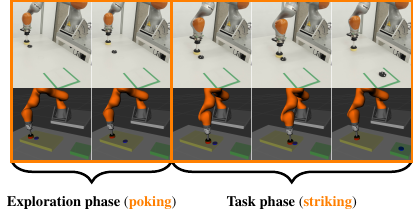}
                \caption{
                Top: The robot's exploration (poking the puck to infer friction and CoM) and task execution (striking to the target). 
                Bottom: RViz view with the workspace (yellow) and target (green). 
                }
            \end{subfigure}
            \vspace{0.5em}
            \begin{subfigure}{\linewidth}
                \includegraphics[width=\linewidth]{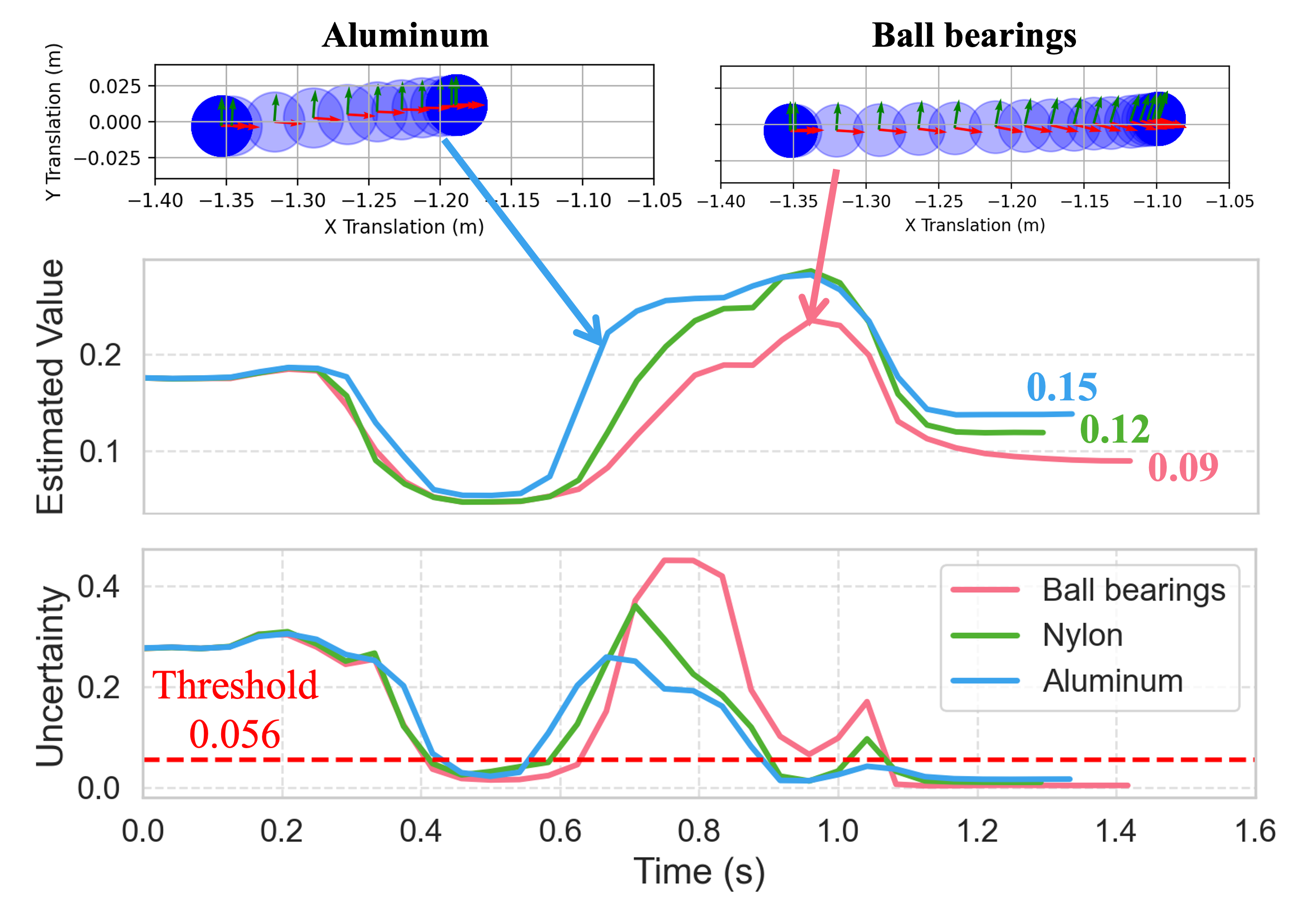}
                \vspace{-1.5em}
                \caption{
                    Estimated dynamic friction and its uncertainty during exploration. 
                    The policy switches when all property uncertainties fall below their thresholds, though we only show dynamic friction here. 
                }
            \end{subfigure}
        \fi
        \vspace{-2em}
        \caption{
            Robot experiments on \texttt{Striking} task. 
        }
        \label{fig:robot_combined}
        \vspace{-4em}
    \end{wrapfigure}
    \subsection{How does our approach perform on a physical robot?}
    \label{s:robot_results}
    
    \paraDraft{Robot experiment setting}
    We deploy our method on the robot, with examples provided in the video and on the~\href{https://marina-aoyama.github.io/poke-and-strike/}{website}. 
    Our approach addresses sim-to-real challenges by learning to explore and estimate task-relevant physical properties of objects, and by enabling the task policy to adjust its motion based on the estimated properties. 
    
    On the \texttt{Striking} task, we evaluated on pucks with three distinct friction properties: ball bearings, nylon, and aluminum. 
    During exploration, each puck exhibits distinct motion patterns due to differences in dynamic friction, as shown in~\cref{fig:robot_combined}. 
    The online property estimator successfully capture these differences, with dynamic friction estimates converging to 0.09 for ball bearings, 0.12 for nylon, and 0.15 for aluminum, as the uncertainty drops. 

    Given the estimated properties, our method achieves 8/9 successful runs across three trials per puck. 
    These results demonstrate that the learned exploration policy and online property estimator provide accurate property estimates, and the uncertainty estimates enable a policy switching, leading to one-shot task success. 
    We provide detailed results for the \texttt{Striking} and \texttt{Edge Pushing} tasks in~\cref{s:appendix_robot}.

\section{Conclusion}
    \label{sec:conclusion}
    We present a task-informed exploration RL approach that enables robots to perform exploratory motions for identifying task-relevant properties.
    Our approach automatically generates task-informed exploration rewards by modeling the sensitivity of a privileged task policy to estimation errors in each property. 
    Additionally, we introduce an uncertainty-based mechanism to transition from exploration to task execution once the property estimates achieve sufficient accuracy for a given task.
    We evaluate our approach on the~\texttt{Striking} and the~\texttt{Edge Pushing} tasks with objects of varying friction and center of mass, demonstrating significantly improved task performance over alternative methods. 
    Our analysis shows that the exploration rewards capture the relative importance of physical properties in two manipulation tasks and the classical CartPole example.
    Finally, we validated our approach in a physical setup, showing that the robot successfully performs exploration followed by task execution using property and uncertainty estimates. 

\clearpage
{\LARGE \bfseries Limitations} 

    \textbf{Correlated properties for exploration reward design.}
    Our method for obtaining the task-informed exploration rewards, by fitting uni-modal functions independently for each property, assumes uncorrelated effects of the properties' estimation errors on the reduction of task performance.
    The correlated case would require designing and testing more complex reward functions. 
    
    \textbf{Geometry and complex dynamics.}
    In this work, we assumed a fixed shaped object. Generalizing over different shapes requires finding suitable input representations for the observed shape when learning the privileged task policy. 
    Extending our approach to represent more complex dynamics—such as non-uniform friction, object deformation, or viscosity, poses the same representation challenge due to the high-dimensionality of the parameter space. 
    Additionally, simulation of such dynamics is computationally expensive, making it demanding for learning policies using RL.
    
    \textbf{Sim2real model mismatch.}
    Despite our method explicitly handling the sim2real transfer by learning to explore and estimate the physical property parameters of the object, we still observed distinct behavior in the highly dynamic striking task when the puck had a shifted center of mass, indicating a significant sim2real mismatch in the dynamic model itself.
    While our method can handle sim-to-real gaps due to parameter mismatch, it is unable to handle mismatches in the dynamic model itself, as it learns parameter estimators in simulation. 

    \textbf{Sensor modalities. }
    Finally, in our experiments, we use only kinematic observations to infer property parameters. 
    It remains unexplored how to best utilize other sensing modalities, such as force-torque, tactile, audio, and vision to identify a wider range of properties and adapt to more diverse tasks and environments.

    

\clearpage
\acknowledgments{
This work is supported by the JST Moonshot R\&D (Grant No. JPMJMS2031). 
We appreciate all reviewers for the valuable feedback. 
}


\bibliography{example}  

\captionsetup{font=normalsize}
\subcaptionsetup{font=normalsize}
\clearpage
\appendix
{\LARGE \bfseries Appendix}

\section{Method}
\subsection{Task-informed exploration framework}
\label{subsec:appendix_framework}   
\begin{figure}[h]
        \centering
        \ifdefined\draftTikz
            \tikzsetnextfilename{fig_diagram}
            \input{fig_diagram.tex}
        \else
            \includegraphics[width=\textwidth]{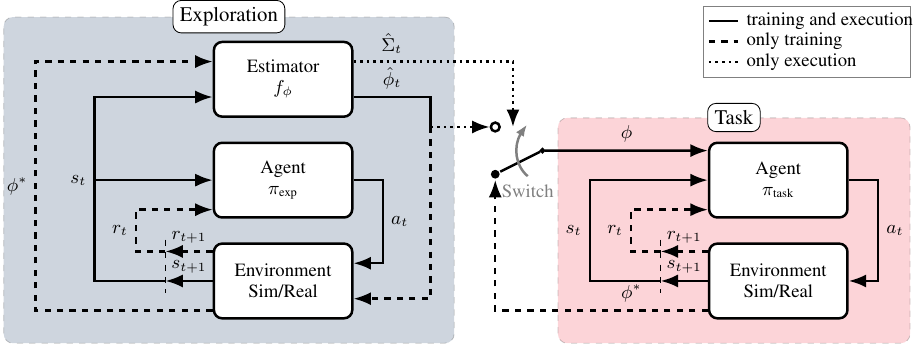}
        \fi
        \caption{%
            Method outline.
            The exploration component involves simultaneous training of an exploration policy~$\pi_{exp}$ and a properties estimator~$f_\phi$, given ground truth property labels~$\phi^\ast$, and executing the policy until the uncertainty~$\hat{\Sigma}_t$ gets lower than a given threshold.
            The task component entails training a task policy, given the privileged information of the ground truth properties~$\phi^\ast$, which during execution takes the last property estimate~$\hat{\phi}$ of the exploration phase as an input.
        }
        \label{fig:method_diagram}
    \end{figure}

\paraDraft{Task-informed exploration framework.}
Our proposed approach utilizes the task policy~$\pi_{task}$ to autonomously discover task-informed exploration rewards for training the exploration policy~$\pi_{exp}$, without manual design. 
These rewards enable us to learn an exploration policy that generates informative motions to estimate the physical properties relevant to the task. 
Additionally, we compute the uncertainty of the physical property estimates and use them as a criterion to switch from the exploration to the task phase, as a surrogate of a estimation error.
The task policy then uses the last estimate of the physical properties~$\hat{\phi}$ from the exploration phase as input.
\cref{fig:method_diagram} outlines the proposed method.

\subsection{Training the physical property and uncertainty estimator}
\label{s:appendix_estimator_loss}
In the property estimator ensemble $f_\phi$, each neural network $i$ outputs a predicted mean $\hat{\phi}_{i,t}$ and covariance $\hat{\Sigma}_{i,t}$, given the current state $s_t$.
To train each network, we model the conditional distribution of the true physical properties as a multivariate Gaussian parameterized by the predicted mean and covariance, i.e. $p(\phi | s_t) = \mathcal{N}(\hat{\phi}_{i,t}, \hat{\Sigma}_{i,t})$.
Therefore, we use the corresponding negative log-likelihood (up to an additive constant) as the loss criterion:
\begin{equation}
    \mathcal{L}_{est} =  \frac{1}{2}\operatorname{ln}\left(\left|\hat{\Sigma}_{i,t}\right|\right)
    + \frac{1}{2}\left(\phi^\ast - \hat{\phi}_{i,t}\right)^{T} \hat{\Sigma}_{i,t}^{-1} \left(\phi^\ast - \hat{\phi}_{i,t}\right).
    \label{e:estimator_loss} 
\end{equation}

\section{Experimental setup}
\label{s:appendix_experimental_details}
\subsection{Tasks}\label{subsec:tasks}   

\begin{figure}[h]
    \centering
    \subfloat[]{%
        \includegraphics[width=0.47\textwidth]{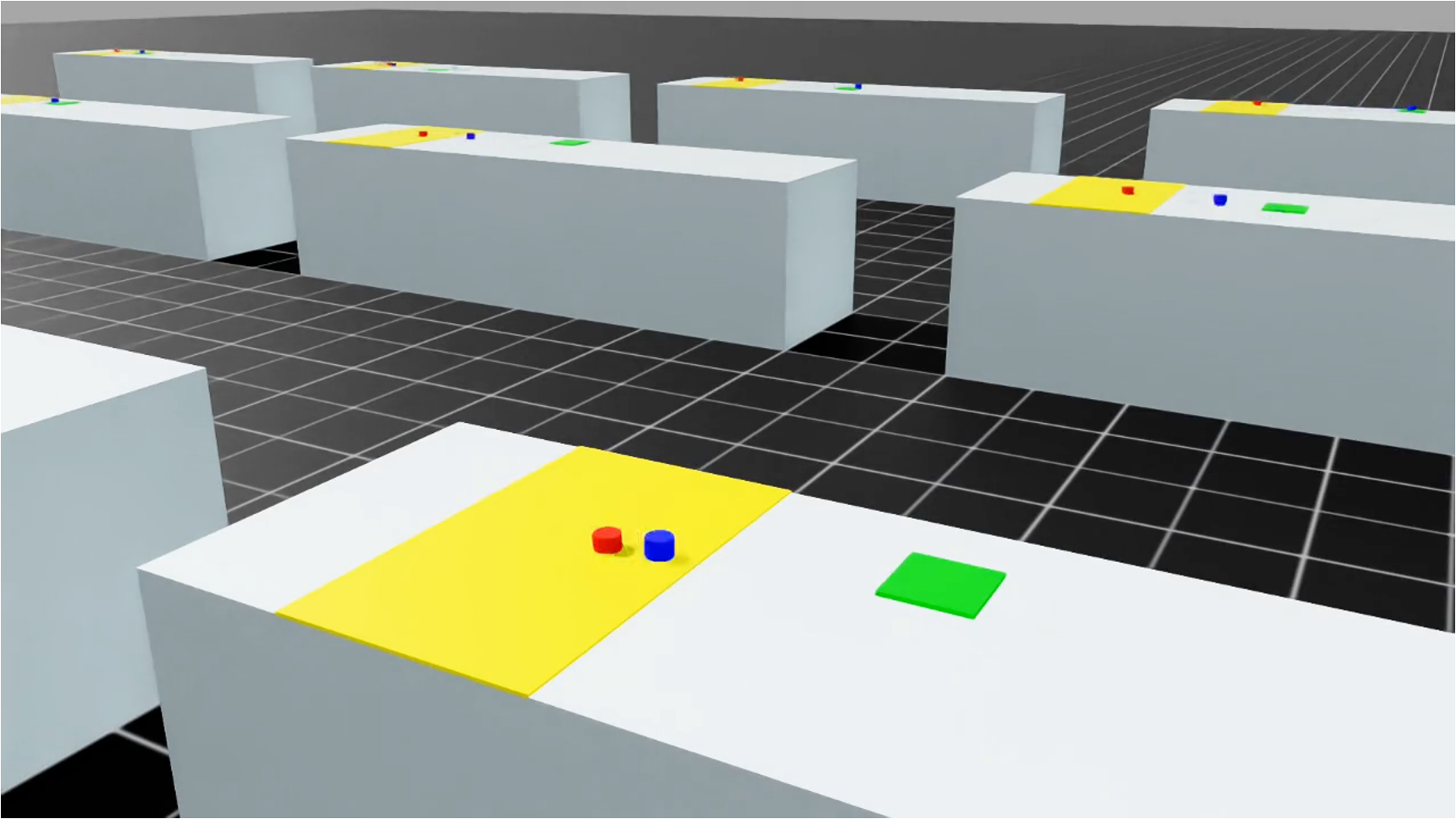}
        \label{fig:sliding_isaac_setup}
    }
    \hfill 
    \subfloat[]{%
        \ifdefined\draftTikz
            \tikzsetnextfilename{fig_sliding}
            \input{fig_sliding.tex}
        \else
            \includegraphics[width=0.47\textwidth]{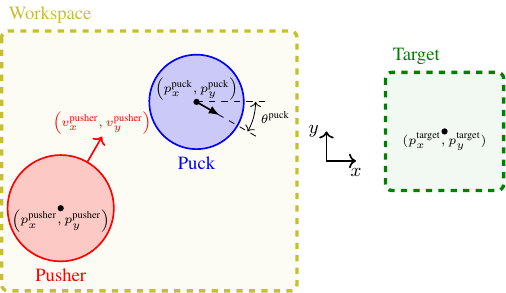}
        \fi
        \label{fig:sliding_schematics}
    }
    \caption{%
        \texttt{Striking} environment 
        (a) in Isaac Lab~\cite{makoviychuk2021isaac, mittal2023orbit} and 
        (b) top view schematics.
        The red cylindrical pusher, on the left of the dark blue puck, moves at a given velocity inside of the yellow reachable workspace.
        The goal of the task is to strike the puck to reach the square green target area.
    }
    \label{fig:sliding}
\end{figure}

\textbf{Striking.} 
The goal of this task is to strike a puck into a given target area beyond the robot's reachable workspace.
\cref{fig:sliding_isaac_setup} shows the simulated environment, and~\cref{fig:sliding_schematics} provides a schematic representation, highlighting the non-overlapping yellow workspace and green target area.
This highly dynamic task exemplifies the scenario where the robot must adjust its motion based on the physical properties of the object and is unable to recover from failure once the puck becomes unreachable, motivating the need to perform exploratory motions before executing the main task.
For each trial, the puck and the robot start from a fixed position, and the task is successful if the puck arrives within a $20\,\si{\centi\meter}\times20\,\si{\centi\meter}$ square at the goal target coordinates~$(p_x^{\text{target}}, p_y^{\text{target}})$. 
The state~$s_t$ consists of the puck's position~$(p_x^{\text{puck}}, p_y^{\text{puck}})$ and orientation~$\theta_{\text{puck}}$, and the robot's pusher position $(p_x^{\text{pusher}}, p_y^{\text{pusher}})$, and the action $a_t$ consists of the pusher velocity~$(v_x^{\text{pusher}}, v_y^{\text{pusher}})$. 
The properties~$\phi$ contain: static friction, dynamic friction, restitution, center of mass (CoM) in the $x$ and $y$ directions, and mass. 

\begin{figure}[h]
    \centering
    \subfloat[]{%
        \includegraphics[width=0.47\textwidth]{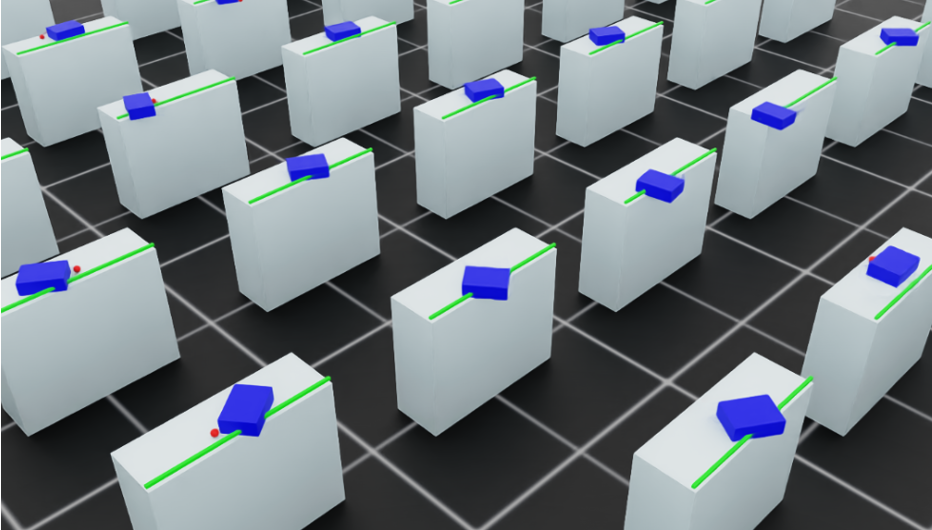}
        \label{fig:edge_pushing_isaac_setup}
    }
    \subfloat[]{%
        \includegraphics[width=0.40\textwidth]{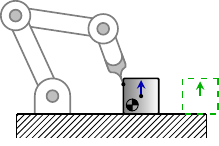}
        \label{fig:edge_pushing_schematics}
    }
    \caption{%
        \texttt{Edge Pushing} environment 
        (a) in Isaac Lab~\cite{makoviychuk2021isaac, mittal2023orbit} and 
        (b) lateral view schematics.
    }
    \label{fig:edge_pushing}
\end{figure}

\textbf{Edge Pushing.} 
This task uses the same setup, randomized physical properties, state, and action definitions as the \texttt{Striking} task.
The goal is to push a box with unknown contents to a target position at the edge of the table. 
Inaccurate estimates of the box’s physical properties can lead to catastrophic and irreversible failure---specifically, the box falling off the table.
\cref{fig:edge_pushing_isaac_setup} shows the simulated environment, and~\cref{fig:edge_pushing_schematics} illustrates the green target position.
The task succeeds if the box arrives within $10,\si{\centi\meter}$ of the target coordinates~$(p_x^{\text{target}}, p_y^{\text{target}})$ without falling off the table. 

\begin{figure}[h]
    \begin{center}
        \subfloat[]{
            \centering
            \includegraphics[width=0.3\textwidth, trim=30mm 0 30mm 0, clip]{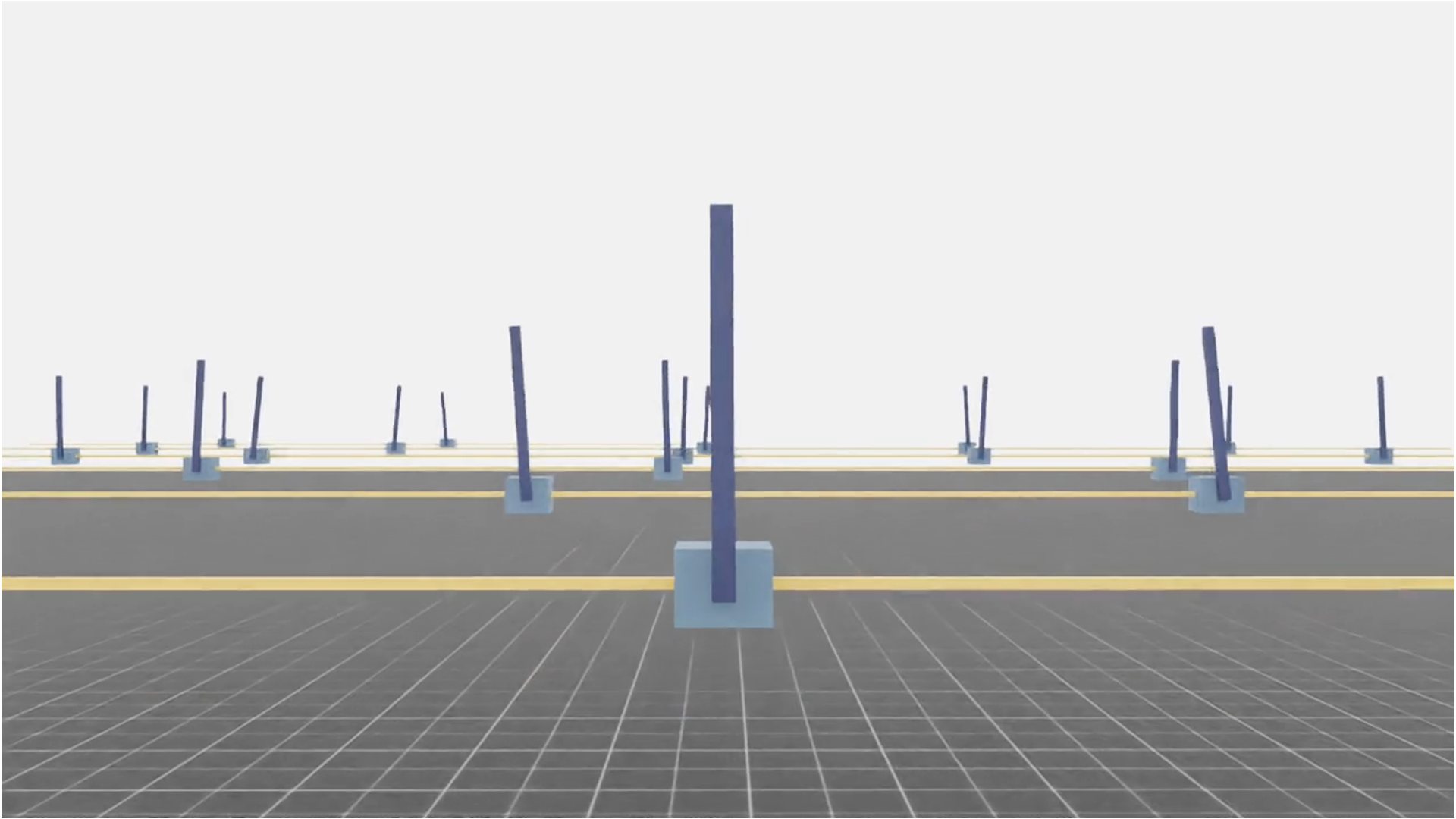}
            \label{fig:cartpole_isaac_sim}
        }
        \subfloat[]{
            \centering
            \ifdefined\draftTikz
                \tikzsetnextfilename{fig_cartpole}
                \input{fig_cartpole.tex}
            \else
                \includegraphics{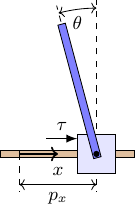}
            \fi
            \label{fig:cartpole_schematics}
        }
    \end{center}
    \caption{%
        \texttt{CartPole} environment 
        (a) in Isaac Lab~\cite{makoviychuk2021isaac, mittal2023orbit} and 
        (b) horizontal view schematics.
        The square cart slides on the horizontal slider when pushed with a force~$\tau$.
        The goal of the task is to balance the rectangular pole vertically, \ie $\theta=0$.
    }
    \label{fig:cartpole}
\end{figure}

\textbf{CartPole.} 
Additionally, we evaluate our approach for modeling task sensitivity to estimation error on the classic CartPole benchmark example from Isaac Lab~\cite{makoviychuk2021isaac} to assess its generalizability to tasks involving a different set of physical properties. 
\cref{fig:cartpole_isaac_sim} shows the simulated environment, and~\cref{fig:cartpole_schematics} provides a schematic representation. 
For each trial, the pole starts from an angle sampled from $\mathcal{U}[-0.25, 0.25]\,\si{\radian}$.
The task is successful if the pole remains upright within a tolerance of $\pi/2\,\si{\radian}$ for 5\,\si{\second}.
The state consists of the current cart position~$p_x$, cart velocity~$\dot{p}_x$, pole angle~$\theta$, and pole angular velocity~$\dot{\theta}$, and  the action $a_t$ specifies the force~$\tau$ applied to the cart along the $x$ direction. 
The properties~$\phi$ contain: cart joint friction, pole joint friction, cart mass, and pole mass. 

\vspace{2em}

\vspace{2em}

\subsection{Simulation setup} \label{sec:simulation_setup}
\paraDraft{Isaac setup}
We develop Striking and CartPole simulation environments using Isaac Lab~\cite{makoviychuk2021isaac, mittal2023orbit}, as depicted in~\cref{fig:sliding_isaac_setup} and ~\cref{fig:cartpole_isaac_sim}. 
We use an NVIDIA GeForce RTX 2080 SUPER GPU and NVIDIA GeForce RTX 4090 throughout the experiments. 

\textbf{Striking.} We abstract the robot as a cylindrical pusher to accelerate training, and model the puck as a fixed-size cylinder. 
We randomize the puck’s physical properties~$\phi$, by sampling from the distributions specified in~\cref{t:combined_randomization_range}.
The environment operates at a control frequency of 24 Hz, with a maximum episode duration of $H=300$ steps or 12.5 seconds. 
Furthermore, we utilize PyBullet~\cite{coumans_pybullet} to evaluate the performance of our uncertainty estimation by rolling out the policy and estimator on a different physics engine from the one used for training. 

\textbf{Edge Pushing.} This environment uses the same setup as \texttt{Striking}, with the puck replaced by a fixed-size box. 

\textbf{CartPole.} We randomize the CartPole properties~$\phi$, by sampling from the distributions specified in~\cref{t:combined_randomization_range}. 
The environment operates at a control frequency of 60 Hz, with a maximum episode duration of~$H=300$ steps or 5 seconds. 

\vspace{1em}
\begin{table}[h]
    \centering
    \setlength{\arrayrulewidth}{0.3mm} 
    \setlength{\tabcolsep}{8pt} 
    \renewcommand{\arraystretch}{1.2} 
    \begin{tabular}{llc}
    \hline\hline
    \textbf{Task} & \textbf{Physical Property} & \textbf{Distribution} \\ \hline
    \multirow{6}{*}{\texttt{Striking}} 
    & Static Friction & $\mathcal{U}[0.05, 0.3]$ \\
    & Dynamic Friction & $\mathcal{U}[0.05, 0.3]$ \\
    & Restitution & $\mathcal{U}[0.0, 1.0]$ \\
    & Mass ($\mathrm{kg}$) & $\mathcal{U}[0.02, 0.5]$ \\
    & Center of Mass Distance ($\mathrm{m}$) & $\mathcal{U}[0, 0.7 \times \text{puck radius}]$ \\
    & Center of Mass Angle ($\mathrm{rad}$) & $\mathcal{U}[0, 2\pi]$ \\ \hline
    \multirow{6}{*}{\texttt{Edge Pushing}} 
    & Static Friction & $\mathcal{U}[0.05, 0.3]$ \\
    & Dynamic Friction & $\mathcal{U}[0.05, 0.3]$ \\
    & Restitution & $\mathcal{U}[0.0, 1.0]$ \\
    & Mass ($\mathrm{kg}$) & $\mathcal{U}[0.02, 0.5]$ \\
    & Center of Mass Distance ($\mathrm{m}$) & $\mathcal{U}[0, 0.7 \times \text{box width}]$ \\
    & Center of Mass Angle ($\mathrm{rad}$) & $\mathcal{U}[0, 2\pi]$ \\ \hline
    \multirow{4}{*}{\texttt{CartPole}} 
    & Cart Joint Friction & $\mathcal{U}[0.0, 1.0]$ \\
    & Pole Joint Friction & $\mathcal{U}[0.0, 1.0]$ \\
    & Cart Mass ($\mathrm{kg}$) & $\mathcal{U}[0.1, 10.0]$ \\
    & Pole Mass ($\mathrm{kg}$) & $\mathcal{U}[0.1, 10.0]$ \\ \hline
    \end{tabular}
    \vspace{0.5em}
    \caption{Randomization range of physical properties. } 
    \label{t:combined_randomization_range}
\end{table}

\subsection{Training setup}\label{sec:training_setup}
\textbf{Privileged task policy.}  
We train the privileged task policy $\pi_{task}$ using Proximal Policy Optimization (PPO)~\cite{schulman2017proximal} with a continuous action space, and stack the last 5 states, $\{s_t, s_{t-1}, \dots, s_{t-4}\}$, to capture temporal information. 
For the \texttt{Striking} and \texttt{Edge Pushing} tasks, the policy function consists of a neural network architecture with four linear layers of sizes 256, 128, 64, and 2, followed by Tanh non-linearities, and the value function consists of the same architecture, but with the final linear layer (1) that outputs the state value prediction.
~\cref{t:ppo_hyperparameters} provides a summary of the key hyperparameters and training settings. 
For the \texttt{CartPole} example we use the default training settings.

\textbf{Task rewards.} 
For the \texttt{Striking} and \texttt{Edge Pushing} tasks, the robot receives a positive reward of $r_t = +30$ for positioning the object (the puck or the box) within the target area and a penalty of $r_t = -15$ if either the object or the robot's end effector violates the workspace boundaries. 
For the \texttt{CartPole} task, we define the reward function as $r_t = +1$ when the pole stays within $\pi/2$ rad of the upright position and the cart remains within the track limits.
Violating either condition results in $r_t = -2$ and immediate episode termination as a failure. 

\textbf{Exploration policy.} 
We also use PPO to train the exploration policy $\pi_{exp}$, consisting of an LSTM layer with 128 hidden units, followed by three fully connected layers with 256, 64, and 2 units, and Tanh non-linearities.
The value function consists of the same architecture, but with the final linear layer (1) that outputs the state value prediction. 
~\cref{t:ppo_hyperparameters} summarizes the key hyperparameters.  

\textbf{Physical property and uncertainty estimator.} 
Each network in the ensemble consists of two LSTM layers with 128 hidden units, followed by a fully connected layer that outputs $2 \times \|\phi\|$ units, where $\|\phi\|$ is the dimensionality of the physical properties being estimated. 
The first $\|\phi\|$ units, to which we apply a Tanh nonlinearity, correspond to the predicted \textit{mean} of the physical properties~$\hat{\phi}_{i,t}$.
The remaining $\|\phi\|$ units correspond to the predicted natural logarithm of the diagonal elements of the \textit{covariance} matrix~$\hat{\Sigma}_{i,t}$.
We use the natural logarithm to improve numerical stability during training and assume a diagonal covariance matrix to simplify parametrisation and reduce the dimensionality of the output.
To form the ensemble, we train five networks with different random initialization weights.

\textbf{Noise.} 
We add two types of noise to the observations: step-wise noise, applied independently at each timestep to simulate sensor noise, and episodic noise, which remains constant throughout the episode to simulate calibration errors. 
~\cref{t:combined_simulation_noise} summarizes the distributions of the observation noise. 

\begin{table}[h]
    \centering
    \setlength{\arrayrulewidth}{0.3mm} 
    \setlength{\tabcolsep}{8pt} 
    \renewcommand{\arraystretch}{1.2} 
    \begin{tabular}{l c}
    \hline\hline
    \textbf{Hyperparameter} & \textbf{Value} \\ \hline
    Parallel Environments & 8192 \\
    Initial Learning Rate (\(\alpha\)) & \(3 \times 10^{-4}\) \\
    Optimizer & Adam \\
    Batch Size & 4096 \\
    Rollout Steps & Task: 16, Exploration: 64 \\
    Number of Epochs & 8 \\
    Discount Factor (\(\gamma\)) & 0.99 \\
    GAE Lambda (\(\lambda\)) & 0.95 \\
    Clip Range (\(\epsilon\)) & 0.2 \\
    Entropy Coefficient (\(\beta\)) & 0.0 \\
    Total Timesteps & 24,000 \\ \hline
    \end{tabular}
    \vspace{0.5em}
    \caption{%
    Key hyperparameters used for PPO training of the task and exploration policies. 
    }
    \label{t:ppo_hyperparameters}
\end{table}
\begin{table}[h]
    \centering
    \setlength{\arrayrulewidth}{0.3mm} 
    \setlength{\tabcolsep}{3pt} 
    \renewcommand{\arraystretch}{1.2} 
    \begin{tabular}{l l c c}
    \hline\hline
    \multirow{2}{*}{\textbf{Task}} & \multirow{2}{*}{\textbf{Input}} & \multicolumn{2}{c}{\textbf{Noise Distribution}} \\ \cline{3-4}
                  &                                 & \textbf{Step}    & \textbf{Episode} \\ \hline
    \multirow{3}{*}{
                    \begin{tabular}{@{}c@{}}
                        \texttt{Striking}/ \\
                        \texttt{Edge Pushing}
                      \end{tabular}
                    } 
                  & Object position (\si{m})               & \( \mathcal{N}(0, 0.0025^2) \) & \( \mathcal{N}(0, 0.0025^2) \) \\
                  & Object orientation (\si{rad})          & \( \mathcal{N}(0, 0.01^2) \)   & \( \mathcal{N}(0, 0.01^2) \)   \\
                  & Pusher position (\si{m})             & \( \mathcal{N}(0, 0.0025^2) \) & \( \mathcal{N}(0, 0.0025^2) \) \\ \hline
    \multirow{4}{*}{\texttt{CartPole}} 
                  & Pole joint position (\si{rad})       & \( \mathcal{N}(0, 0.15^2) \)    & \( \mathcal{N}(0, 0.15^2) \) \\
                  & Pole joint velocity (\si{rad/s})     & \( \mathcal{N}(0, 0.15^2) \)    & \( \mathcal{N}(0, 0.15^2) \)   \\
                  & Cart joint position (\si{m})         & \( \mathcal{N}(0, 0.05^2) \)    & \( \mathcal{N}(0, 0.05^2) \) \\
                  & Cart joint velocity (\si{m/s})       & \( \mathcal{N}(0, 0.05^2) \)    & \( \mathcal{N}(0, 0.05^2) \) \\ \hline
    \end{tabular}
    \vspace{0.5em}
    \caption{%
    Summary of step-wise noise, simulating sensor noise at each timestep, and episodic noise, simulating calibration and systematic errors across the entire episode, for the Striking and CartPole tasks.}
    \label{t:combined_simulation_noise}
\end{table}

\clearpage
\subsection{Hardware setup}\label{sec:hardware_setup}
\begin{figure}[h]
    \centering
    \ifdefined\draftTikz
        \tikzsetnextfilename{fig_coverimage}
        \input{fig_coverimage.tex}
    \else
        \includegraphics{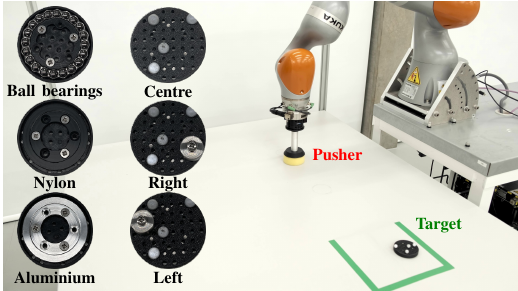}
    \fi
    \caption{%
        The \textbf{task-informed exploration} approach enables the robot to autonomously learn how to explore and identify the physical properties of objects relevant to a given task.
        For one-shot tasks such as striking, the robot must first identify the object's properties through exploratory motions to achieve the task success. 
    }
    \label{f:hardware_img}
\end{figure}

\paraDraft{Kuka setup}
We evaluate our approach on a physical setup using a KUKA iiwa robot arm equipped with a pusher extension, as shown in~\cref{f:hardware_img}. 
We use a Vicon motion capture system to track the object's current pose, and map the policy actions (\ie pusher velocities) to robot joint configurations using inverse kinematics. 
We evaluate two hardware tasks: \texttt{Striking} a puck with three friction levels and three center-of-mass locations (\cref{f:hardware_img}), and \texttt{Edge Pushing} a box of eggs placed on one of two sides (\cref{f:robot_tasks}). 


\section{Additional results} \label{sec:additional_results} 
\subsection{Exploration policy training --- Striking}
\label{s:exploration_results}
\begin{figure}[h]
    \centering
    \includegraphics[width=0.7\textwidth]{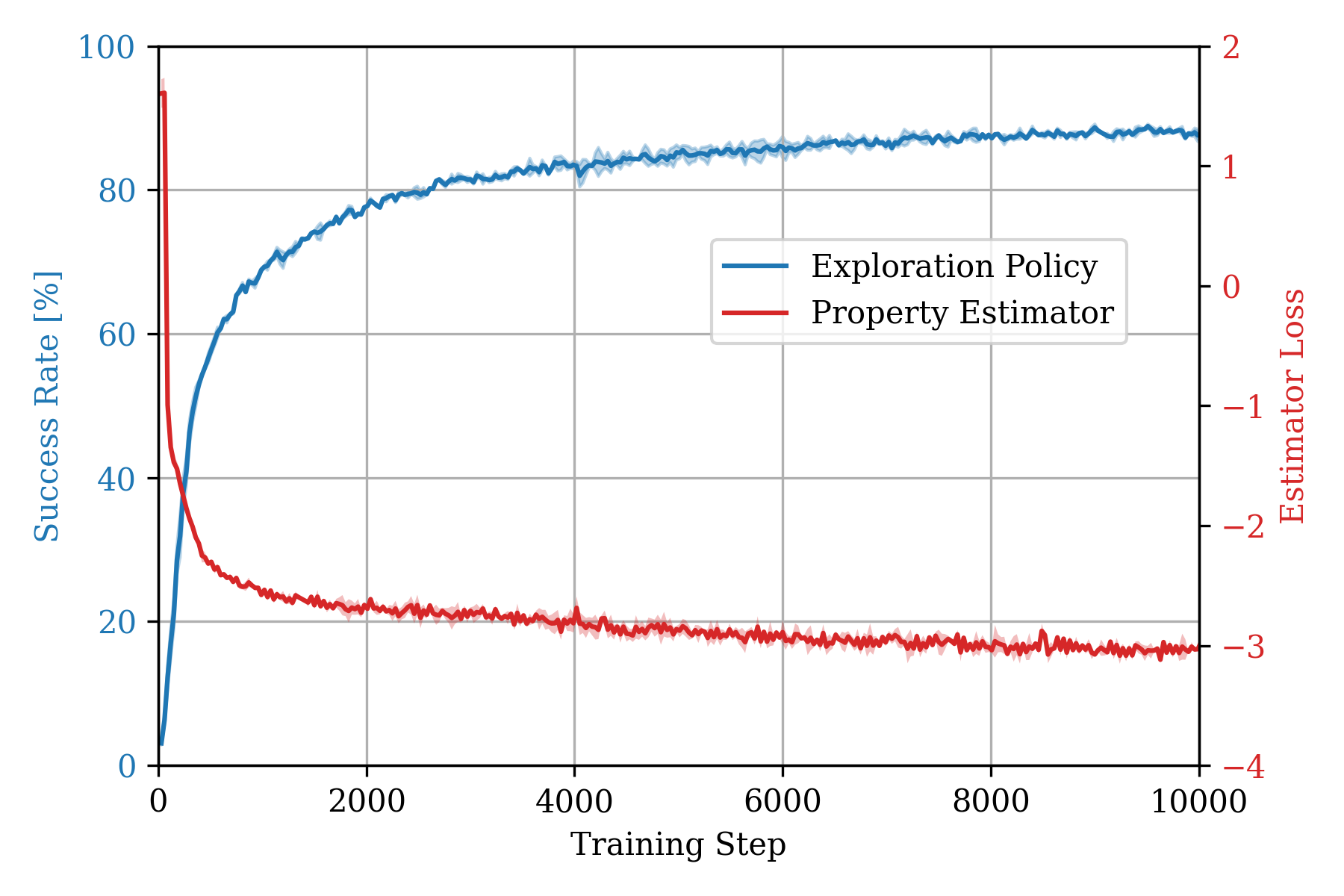}
    \caption{
    Exploration policy training and estimator loss of LSTM-based estimator. 
    }
    \label{f:exploration_training_curves}
\end{figure}

\textbf{Simultaneous training of exploration policy and property estimator. }
\cref{f:exploration_training_curves} shows the training performance of the exploration policy $\pi_{\text{exp}}$, trained using task-informed exploration rewards and the estimator loss defined in~\cref{e:estimator_loss}, for the striking task. 
The results show the average over three random seeds, with shaded regions indicating the standard deviation. 
We demonstrate that, despite the exploration rewards being non-stationary due to their dependence on the simultaneously trained estimator, on-policy RL achieves stable training and consistently converges to an exploration success rate above 90\% across seeds. 

\begin{figure}[h]
    \centering
    \includegraphics[width=0.7\textwidth]{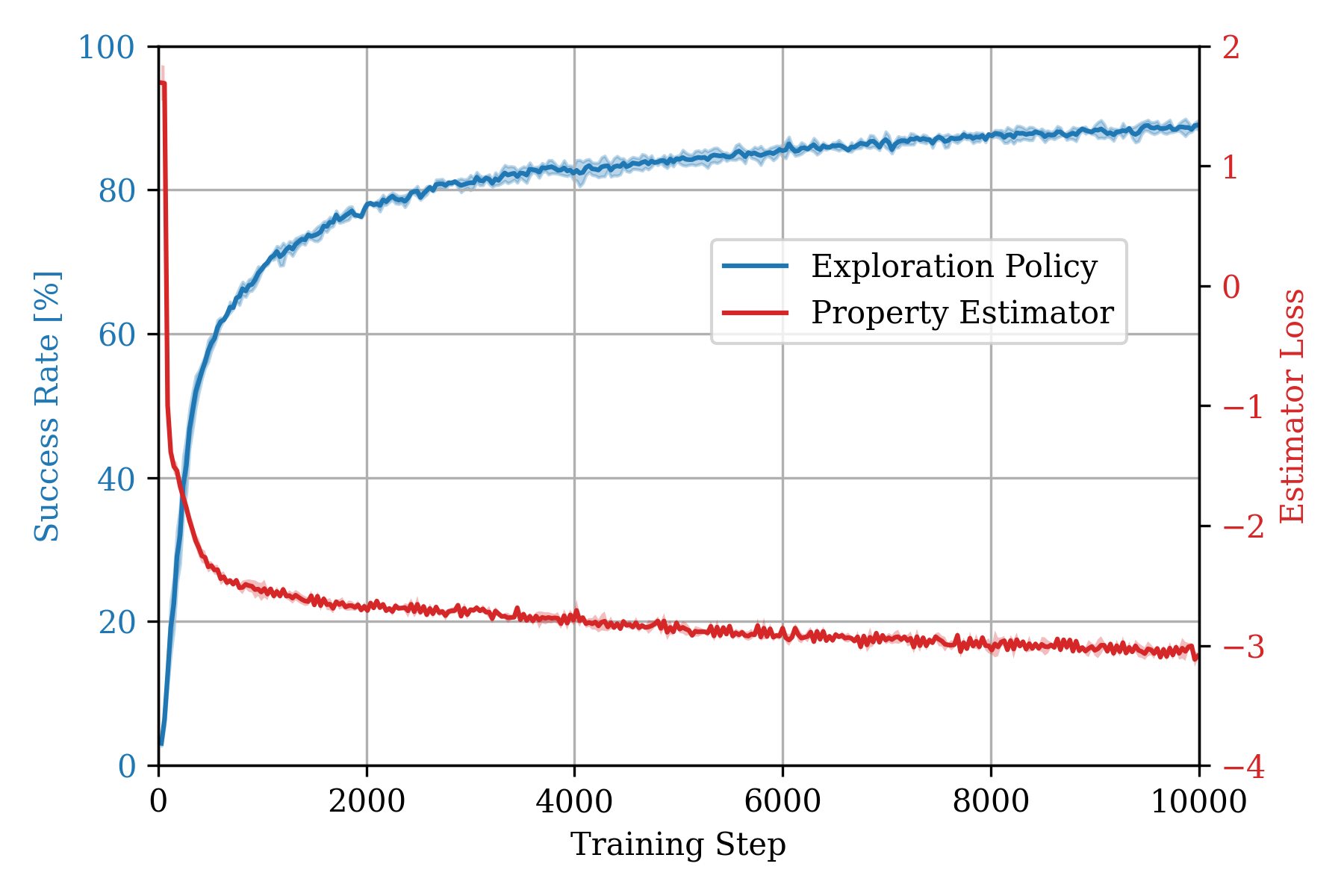}
    \caption{
    Exploration policy training and estimator loss of Transformer-based estimator. 
    }
    \label{f:exploration_training_curves_transformer}
\end{figure}

\textbf{Alternative system identification approaches. }
Our framework supports alternative system identification modules, as long as they can output both estimates and uncertainty. 
While the LSTM-based property estimator is sufficient for our short exploration horizon, we also tested a Transformer-based estimator, achieving similar exploration success (\cref{f:exploration_training_curves_transformer}). 
Our method can accommodate such models for tasks requiring longer temporal dependencies during exploration or multi-modal inputs. 
Additionally, although we chose a learned property estimator for fast, online inference and ease of integration into an RL pipeline without additional simulation queries, investigating alternative modules (\eg sim-in-the-loop approaches) is promising future work. 

\clearpage
\subsection{Edge Pushing baseline comparison  --- Edge Pushing}
\label{s:edgepushing_baseline_comparison}
\begin{figure}[h]
    \centering
    \includegraphics[width=0.7\textwidth]{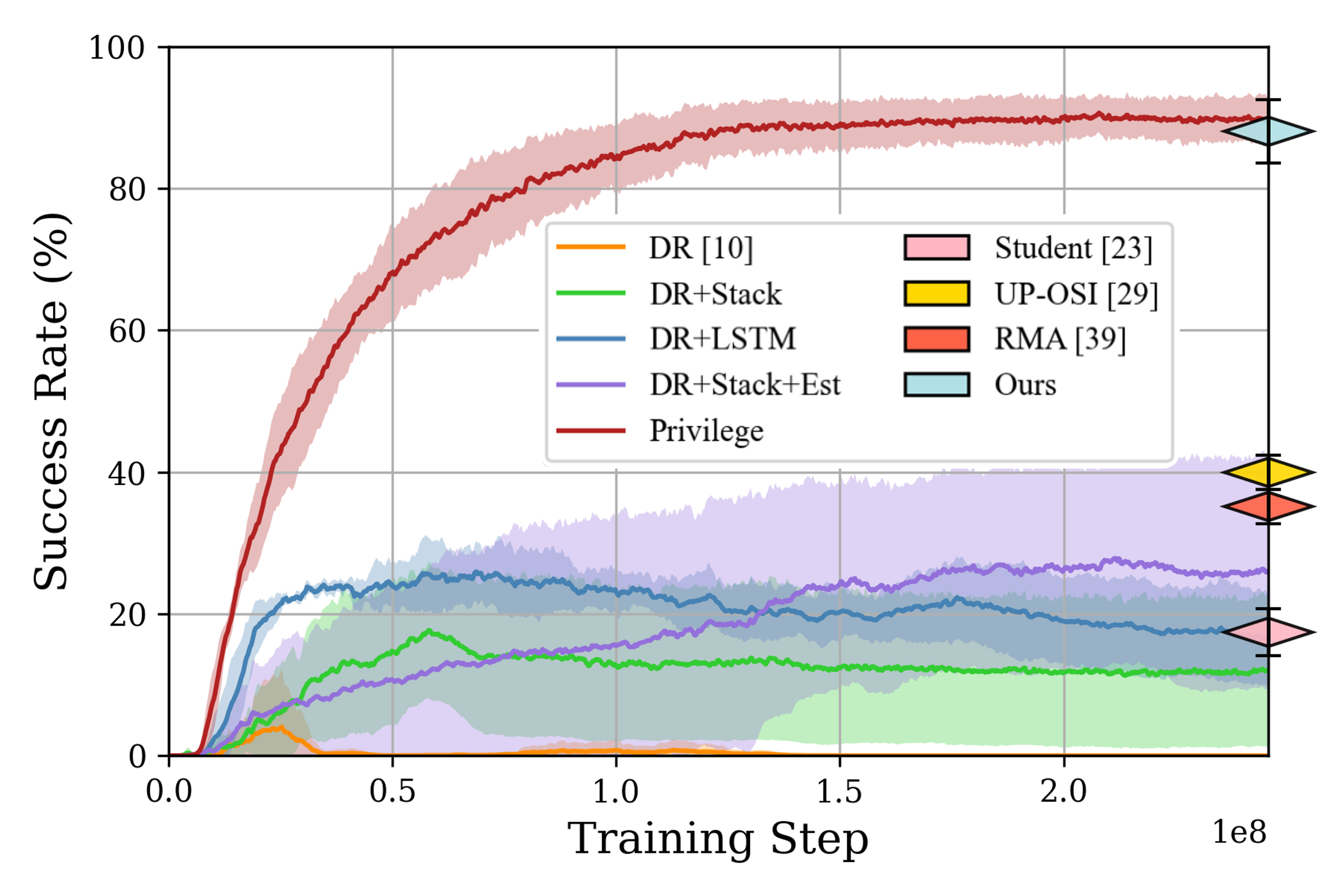}
    \caption{
    Performance with different training configurations of the control policy for Edge Pushing task. 
    We report mean and standard deviation across five training seeds. 
    }
    \label{f:edgepushing_training_curves}
\end{figure}

\subsection{Task-agnostic vs. task-informed exploration rewards} 
\label{s:appendix_ablation_table}
    \begin{table}[h]
        \centering
        \setlength{\arrayrulewidth}{0.2mm}
        \setlength{\tabcolsep}{3pt}
        \renewcommand{\arraystretch}{1.2}
        \begin{tabular}{l c c}
            \noalign{\hrule height 0.3mm}
            \multirow{2}{*}{Method} & \multicolumn{2}{c}{Success Rate (\%)} \\ 
            & Exploration & Task         \\ \hline
            Task-agnostic Min.  & 8.0   & 2.9 \\
            Task-agnostic Max.  & 91.7  & 47.3 \\
            \textbf{Task-informed}          & 92.3  & 90.1\\
            \noalign{\hrule height 0.3mm}
        \end{tabular}
        \vspace{1em}
        \caption{%
            Ablation on task-agnostic vs. task-informed exploration rewards. 
            We provide detailed results in~\cref{s:appendix_exp_rewards_results}. 
        }
        \label{t:threshold_comparison}
    \end{table}

\clearpage
\subsection{Sensitivity analysis --- Striking, Edge Pushing, CartPole}
\label{s:appendix_sensitivity}
While our approaches can apply any uni-modal functions, we applied three different uni-modal distributions, the Gaussian, Beta, and Gamma distributions, to model the sensitivity of each property to each task. 
    \begin{figure}[h]
        \centering
        \subcaption*{\texttt{Striking}}

        \vspace{1mm}

        \begin{minipage}{0.7\textwidth}
            \centering
            \includegraphics[width=\linewidth]{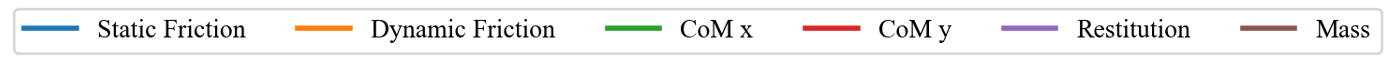} 
        \end{minipage}
        
        \begin{subfigure}[b]{0.32\textwidth}
            \includegraphics[width=\linewidth]{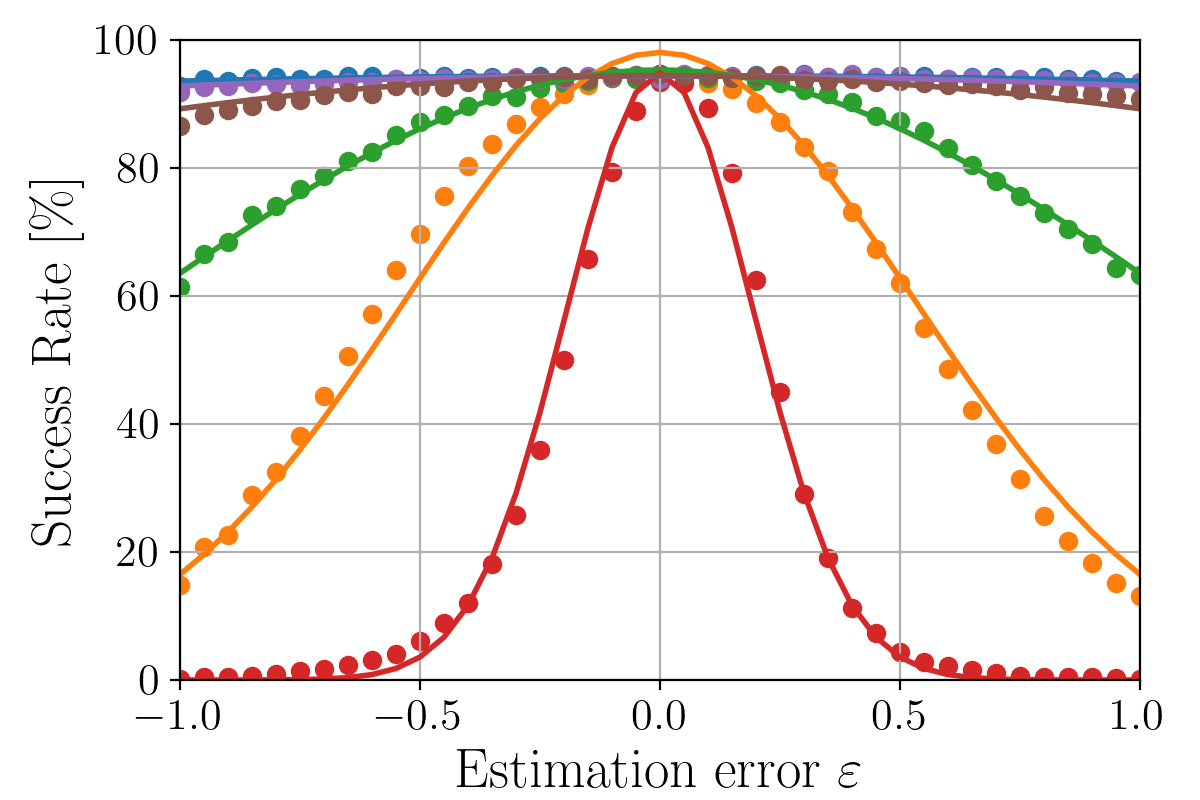}
            \caption{Gaussian function}
            \label{fig:subfig1}
        \end{subfigure}
        \hspace{1mm}
        \begin{subfigure}[b]{0.32\textwidth}
            \includegraphics[width=\linewidth]{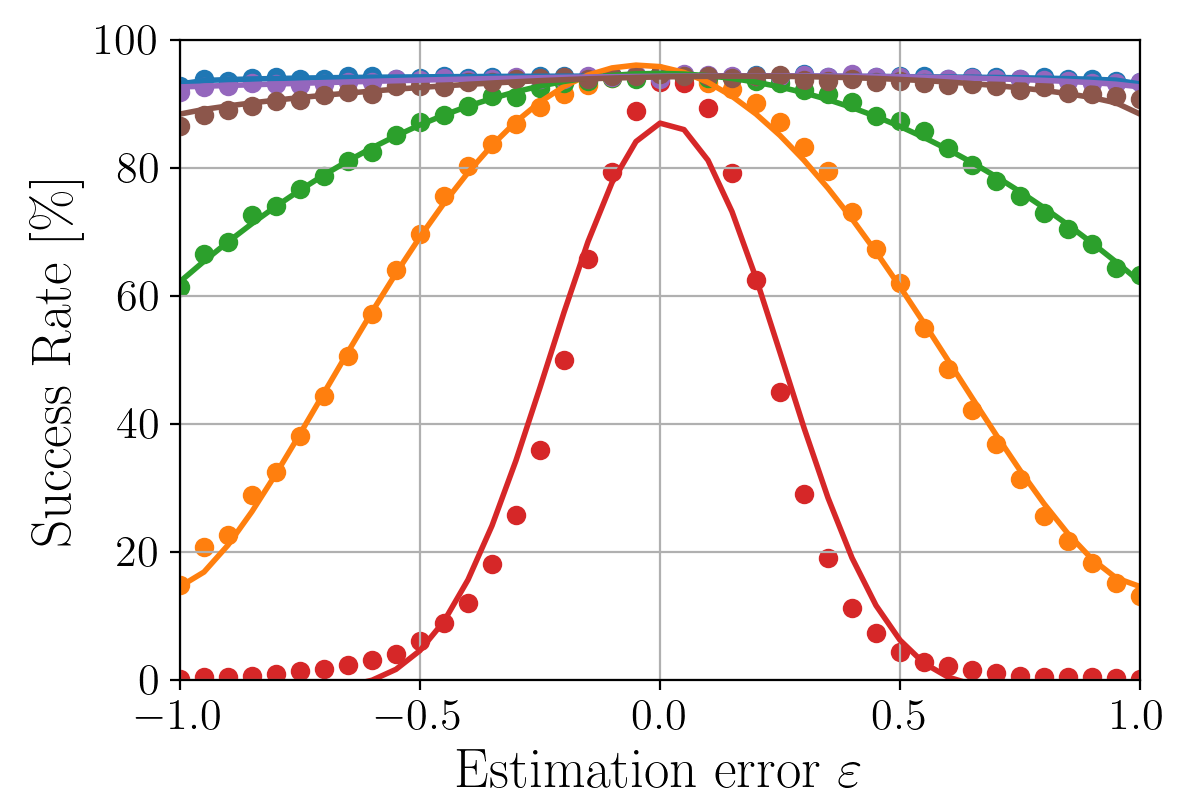}
            \caption{Beta function}
            \label{fig:subfig2}
        \end{subfigure}
        \hspace{1mm}
        \begin{subfigure}[b]{0.32\textwidth}
            \includegraphics[width=\linewidth]{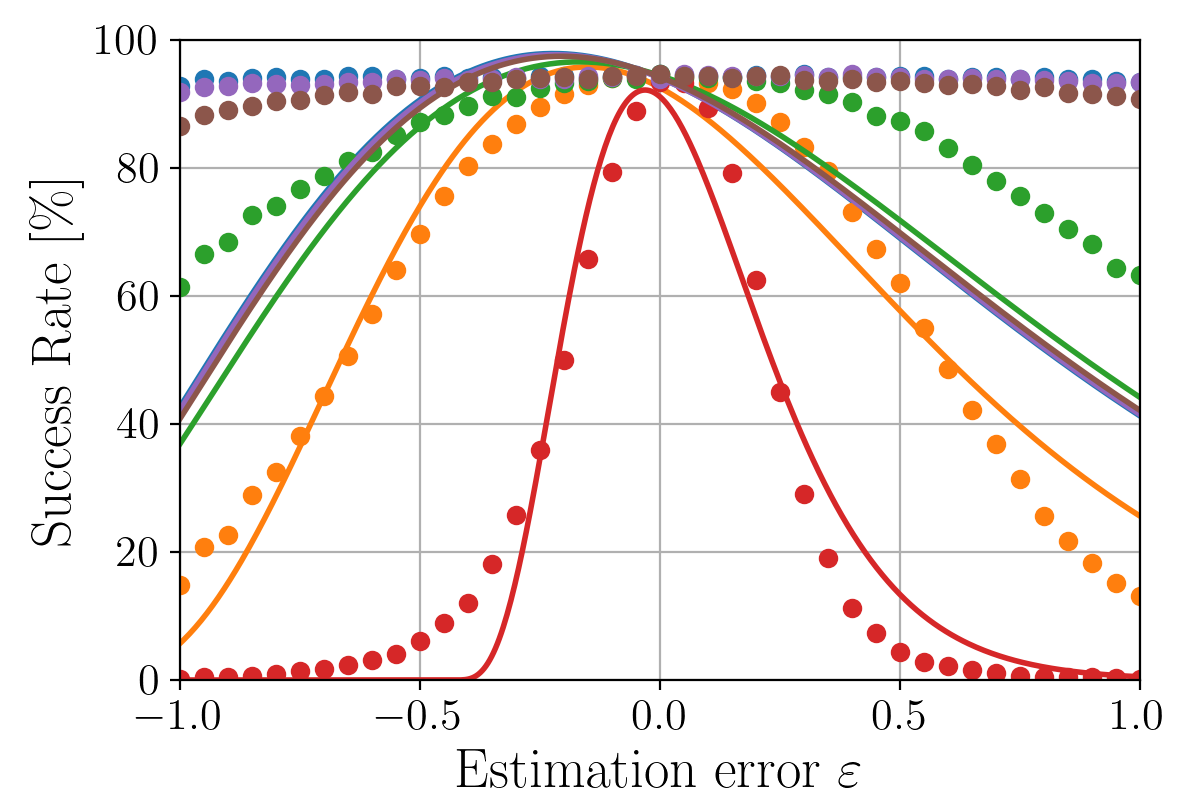}
            \caption{Gamma function}
            \label{fig:subfig3}
        \end{subfigure}
    
        \vspace{3mm}  
    
        \subcaption*{\texttt{Edge Pushing}}

        \vspace{1mm}

        \begin{minipage}{0.7\textwidth}
            \centering
            \includegraphics[width=\linewidth]{img/manipulation_unimodal_legend_cropped.png} 
        \end{minipage}
        
        \begin{subfigure}[b]{0.32\textwidth}
            \includegraphics[width=\linewidth]{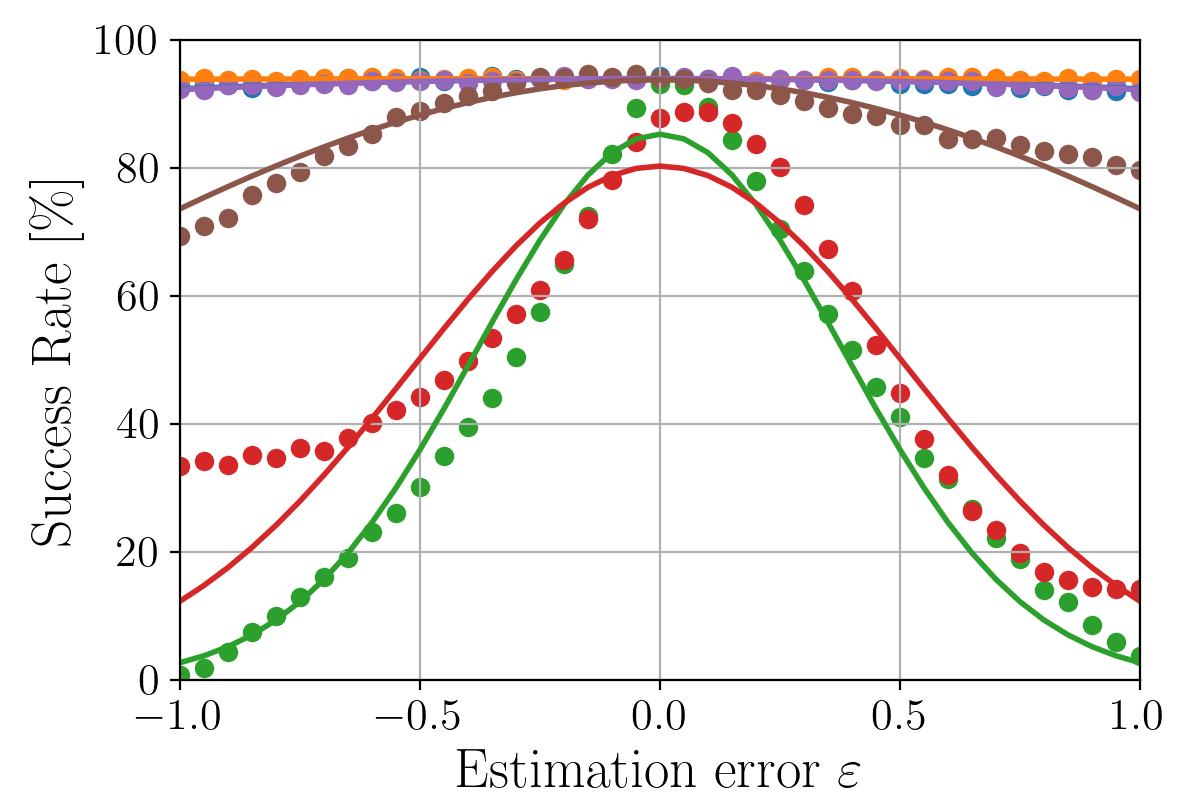}
            \caption{Gaussian function}
            \label{fig:subfig7}
        \end{subfigure}
        \hspace{1mm}
        \begin{subfigure}[b]{0.32\textwidth}
            \includegraphics[width=\linewidth]{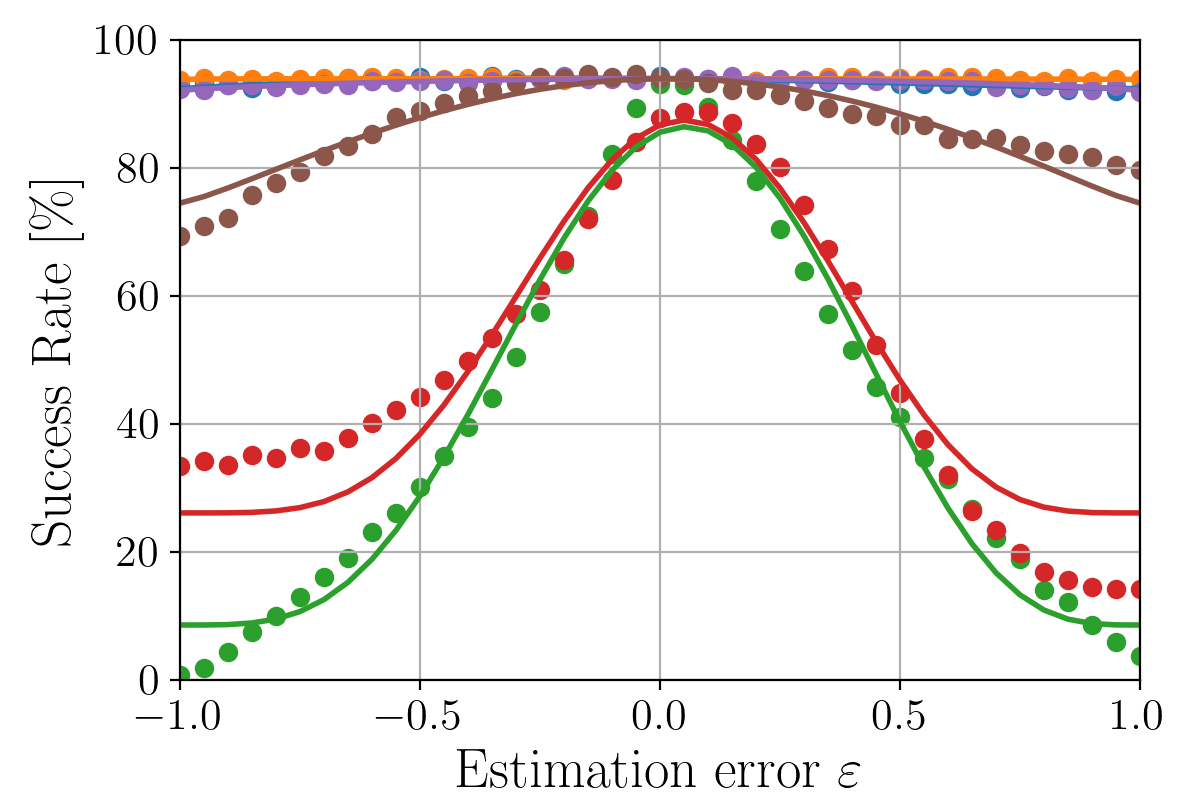}
            \caption{Beta function}
            \label{fig:subfig8}
        \end{subfigure}
        \hspace{1mm}
        \begin{subfigure}[b]{0.32\textwidth}
            \includegraphics[width=\linewidth]{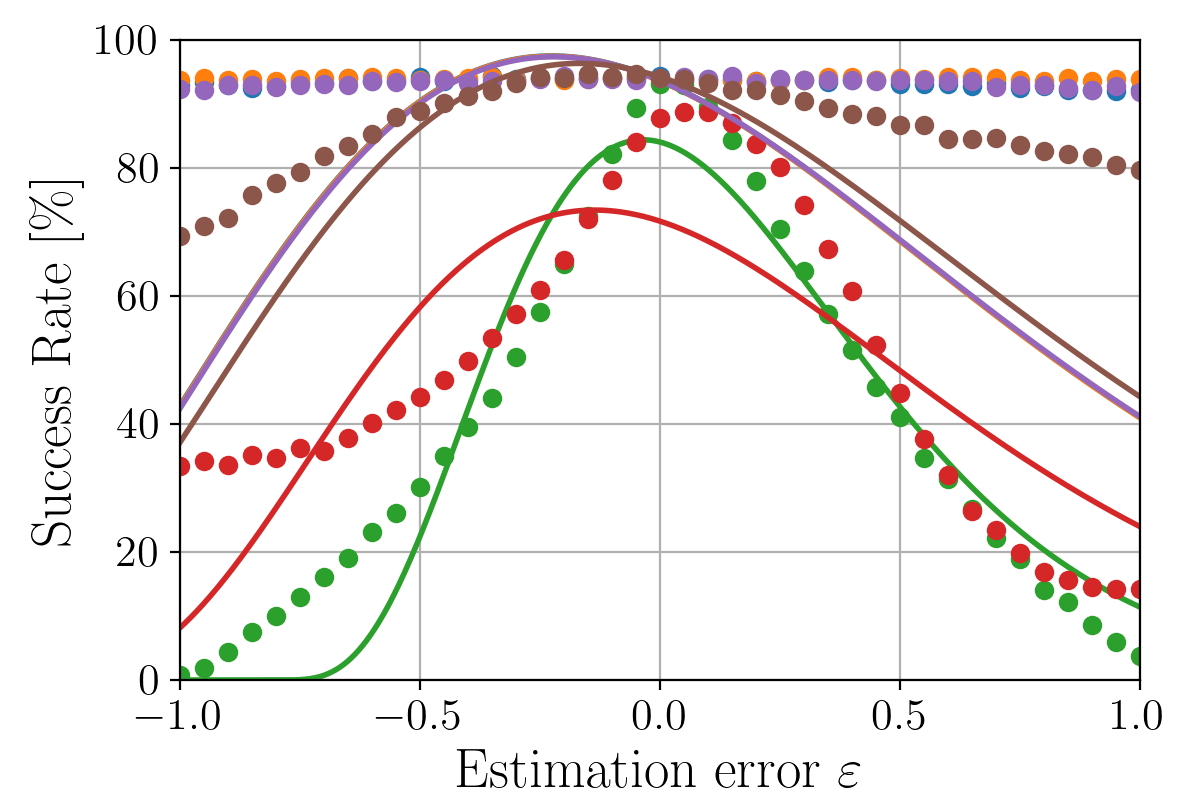}
            \caption{Gamma function}
            \label{fig:subfig9}
        \end{subfigure}
    
        \vspace{3mm}  
    
        \subcaption*{\texttt{CartPole}}

        \vspace{1mm}

        \begin{minipage}{0.5\textwidth}
            \centering
            \includegraphics[width=\linewidth]{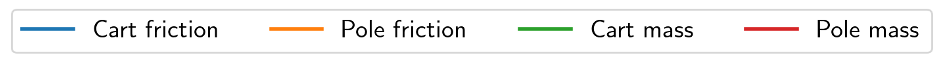} 
        \end{minipage}
        
        \begin{subfigure}[b]{0.32\textwidth}
            \includegraphics[width=\linewidth]{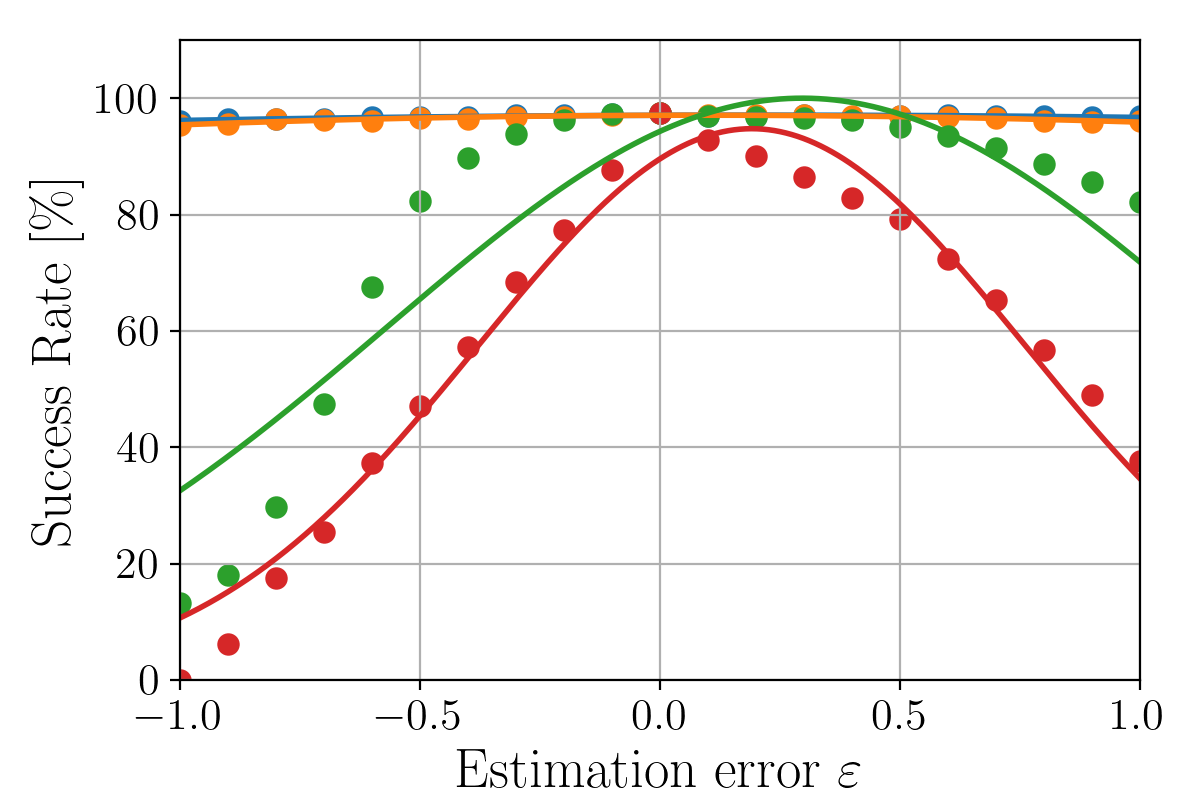}
            \caption{Gaussian function}
            \label{fig:subfig4}
        \end{subfigure}
        \hspace{1mm}
        \begin{subfigure}[b]{0.32\textwidth}
            \includegraphics[width=\linewidth]{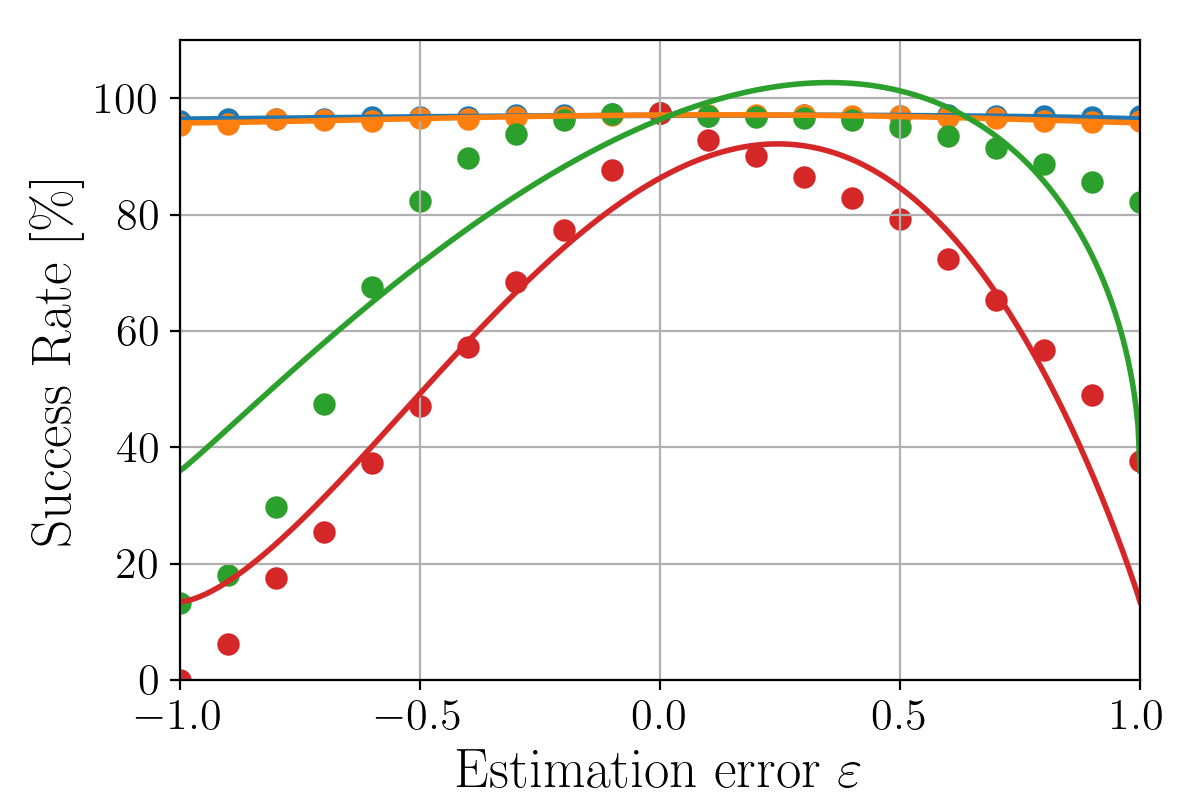}
            \caption{Beta function}
            \label{fig:subfig5}
        \end{subfigure}
        \hspace{1mm}
        \begin{subfigure}[b]{0.32\textwidth}
            \includegraphics[width=\linewidth]{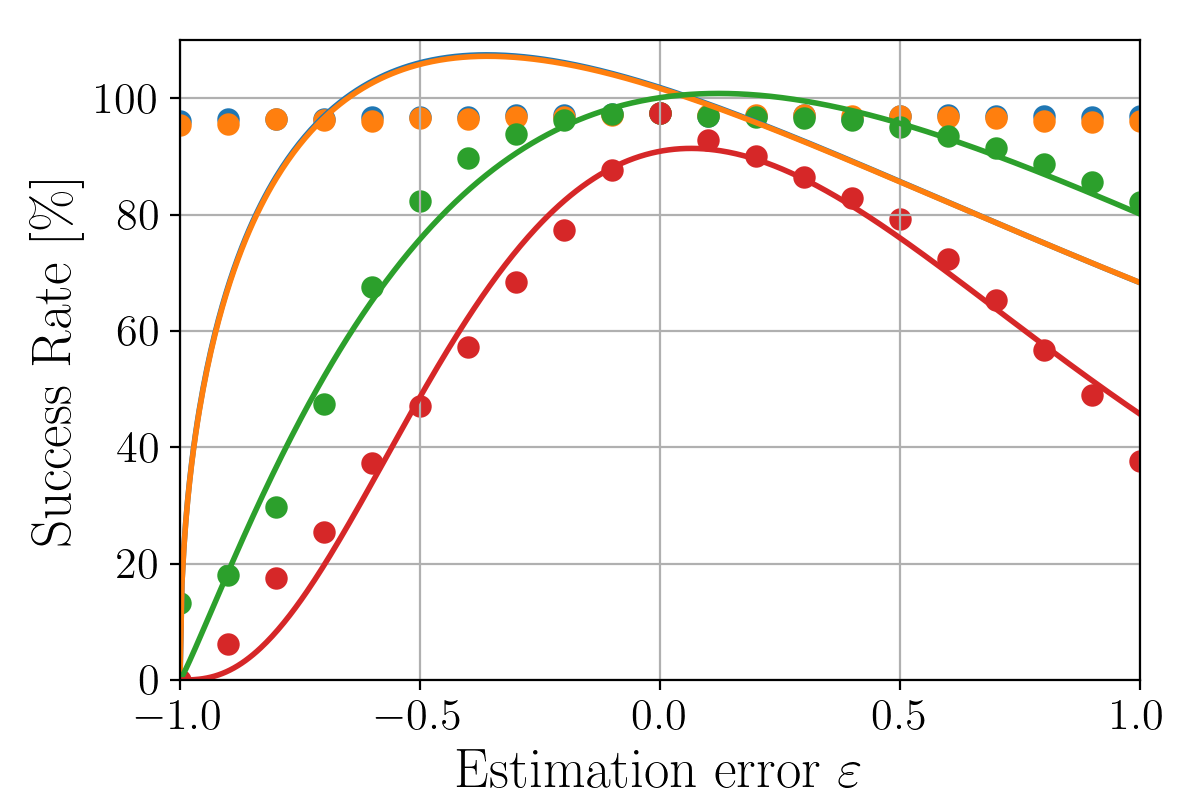}
            \caption{Gamma function}
            \label{fig:subfig6}
        \end{subfigure}
    
        \caption{Different uni-modal functions fitted to model the sensitivity of the task success rate to the estimation error on each property for the \texttt{Striking}, \texttt{Edge Pushing}, and \texttt{CartPole} tasks.}
        \label{fig:combined_uni-modal}
    \end{figure}

\clearpage
\subsection{Estimation error during exploration policy training} 
\label{s:appendix_exp_rewards_results}
\begin{figure}[h]
        \centering
        \ifdefined\draftTikz
            \tikzsetnextfilename{fig_exploration_policy_learning}
            \input{fig_exploration_policy_learning.tex}
        \else
            \includegraphics[width=0.8\linewidth]{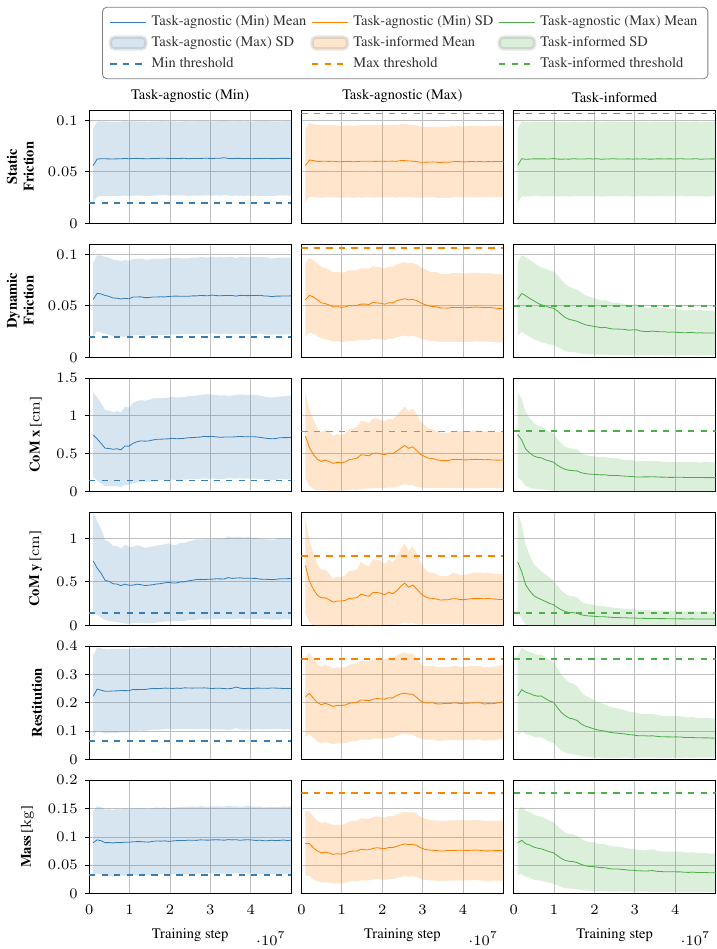}
        \fi
        \caption{%
            RMSE of the property parameter estimation at the end of the episode during training (solid line) and the estimation threshold (dashed line).
        }
        \label{f:estimator_training}
    \end{figure}

~\cref{f:estimator_training} illustrates how the Root Mean Squared Error (RMSE) of the property estimates at the end of the exploration episode evolves during training. 
We report mean and standard deviation (SD) across the training environments.
With the task-informed exploration policy, the estimation error for relevant physical properties, such as CoM and dynamic friction, decreases, while the error for less relevant properties remains high. 
By the end of training, the estimation errors for all properties fall below the thresholds indicated by the dotted lines. 
In contrast, task-agnostic exploration policies result in high estimation errors across all physical properties. 
The exploration reward with high estimation thresholds for all properties (Task-agnostic (Max.)) proves non-informative. 
Due to the high thresholds, even large estimation errors fall below them, failing to incentivize accurate property estimation.
On the other hand, setting thresholds too low for all properties (Task-agnostic (Min.)) makes it difficult to achieve the required thresholds, leading to sparse rewards. 
Additionally, training becomes more challenging, as it requires learning more complex behaviors to accurately estimate all properties.

\clearpage
\subsection{Can we use uncertainty for policy switching?}
    \label{s:uncertainty_results}
\begin{figure}[h]
        \centering
        \ifdefined\draftTikz
            \tikzsetnextfilename{fig_error_uncertainty}
            \input{fig_error_uncertainty.tex}
        \else
            \includegraphics{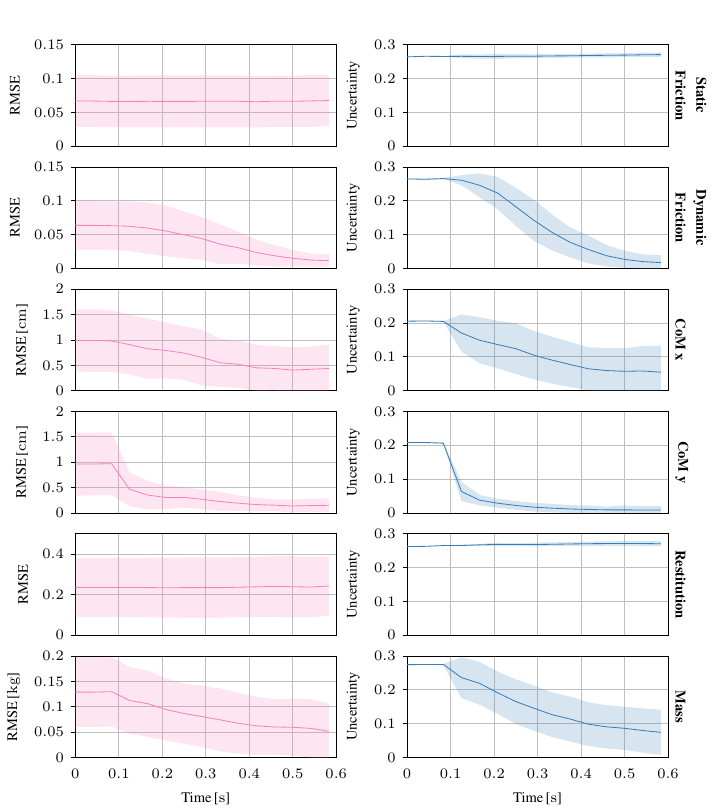}
        \fi
        \caption{%
            Estimation errors and uncertainties of property parameters during the exploration episode.
            As the exploration progresses, both the estimation errors and uncertainties decrease for task-relevant properties.
        }
        \label{f:uncertainty_plot_single_episode}
    \end{figure}

    \textbf{Relationship between uncertainty and estimation error.} 
    First, we evaluate whether the estimated uncertainty reflects the estimation errors by rolling out the exploration policy 100 times in Isaac Lab. 
    ~\cref{f:uncertainty_plot_single_episode} shows the mean and standard deviation of the estimation errors and uncertainties over the first 14 timesteps ($\approx 0.6$ seconds) of the exploration episodes. 
    The plots demonstrate that as the exploration progresses, both the estimation errors and uncertainties decrease for task-relevant properties, such as the CoM in the y-direction and dynamic friction.
    This result suggests that the uncertainty estimates effectively reflect the estimation error, supporting our approach of using uncertainty as a surrogate when the estimation error is inaccessible at test time. 
    
\begin{figure}[h]
        \centering
        \includegraphics[width=0.98\textwidth]{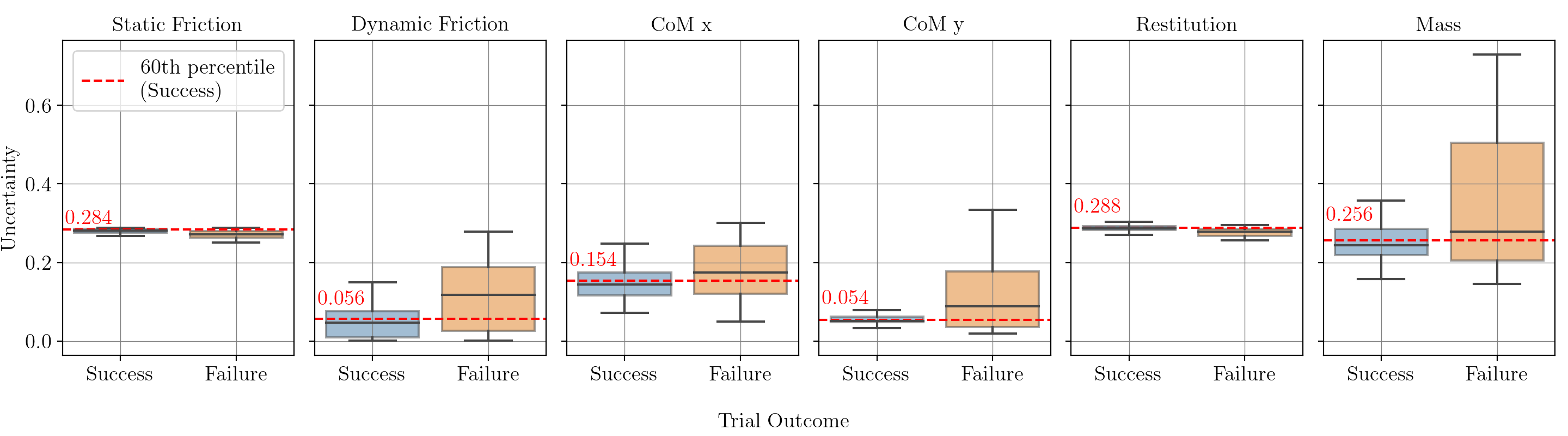}
        \caption{
        Relationship between uncertainties and task outcomes when rolling out the exploration and task policies in PyBullet. 
        Low uncertainties lead to task success, while high uncertainties correlate with failures, particularly for task-relevant properties such as CoM in the y-direction and dynamic friction.
        }
        \label{f:uncertainty_boxplot}
    \end{figure}

    \textbf{Relationship between uncertainty and task success.} 
    Next, we evaluate the relationship between uncertainty estimates and task success. 
    For this evaluation, we roll out the task policy 100 times in PyBullet using estimates from the exploration policy, running for the maximum episode length.
    We use a simulator different from the one used for training to better assess uncertainty when there is a domain gap between training and testing, as we intend to use these uncertainties in a physical setup. 
    ~\cref{f:uncertainty_boxplot} presents box plots of uncertainty at the end of the exploration episode for successful and failed trials. 
    The plots show that low uncertainty in task-relevant property estimates leads to task success, while high uncertainty results in failure. 
    These findings suggest that uncertainty estimates during exploration predicts task outcomes and indicates when property estimates are sufficiently accurate to transition from exploration to the task phase. 

\section{Robot experiments} 
\label{s:appendix_robot}
\subsection{Striking} 
\begin{figure}[h]
        \centering
        \ifdefined\draftTikz
            \tikzsetnextfilename{fig_robot_exp_friction}
            \input{fig_robot_exp_friction.tex}
        \else
            \includegraphics{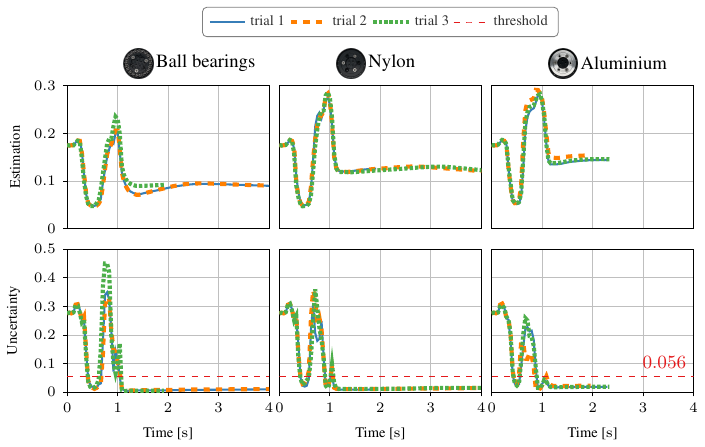}
        \fi
        \caption{%
            Estimates of \textbf{dynamic friction} and its associated uncertainty during exploration policy execution on the physical setup. 
            Dynamic friction estimates consistently converge to 0.9, 0.12, and 0.15 for the pucks with ball bearings, nylon, and aluminum, respectively. 
        }
        \label{fig:robot_estimation_friction}
        \vspace{1em}
    \end{figure}

\textbf{Varying surface friction.} 
We provide dynamic friction estimates and its uncertainty for all trials on pucks with different levels of friction. 
~\cref{fig:robot_estimation_friction} shows that the dynamic friction estimates consistently converge to 0.09 for ball bearings, 0.12 for nylon, and 0.15 for aluminum, with uncertainty dropping below 0.056, the uncertainty thresholds computed in~\cref{f:uncertainty_boxplot}. 

\begin{figure}[h]
        \centering
        \ifdefined\draftTikz
            \tikzsetnextfilename{fig_robot_exp_com}
            \input{fig_robot_exp_com.tex}
        \else
            \includegraphics{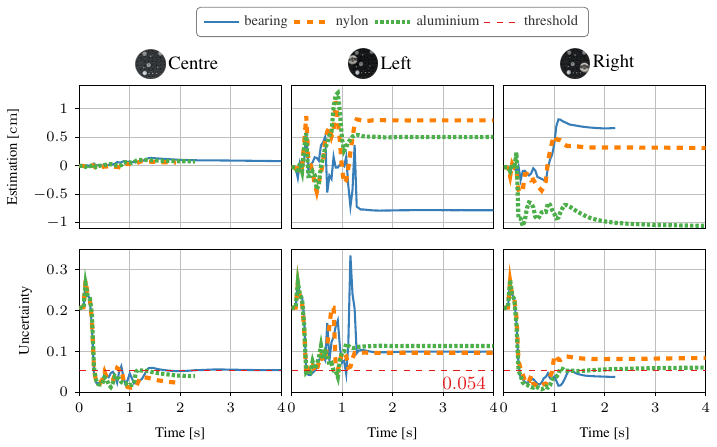}
        \fi
        \caption{%
            Estimates of the~$y$ component of the \textbf{center of mass} and its associated uncertainty during exploration policy execution on the physical setup.
            Uncertainty decreases in successful trials, while it remains high (\ie above the threshold~$0.054$ obtained from~\cref{f:uncertainty_boxplot}) in failed trials.
        }
        \label{fig:robot_estimation_com}
    \end{figure}

    \textbf{Shifted center of mass.} We also evaluated pucks with three varying locations of the center of mass, as shown in the second column in~\cref{f:hardware_img}. 
    When the CoM is in the center, the robot successfully estimates its location and completes the task, achieving 3/3 successful runs. 
    However, for pucks with shifted CoM—either to the left or right—exploration fails due to inaccurate estimates and consistently high uncertainty, as shown in~\cref{fig:robot_estimation_com}. 
    We observe that, while shifts in the CoM primarily cause spinning with minimal trajectory deviation in the Isaac Lab training environment, they result in more pronounced trajectory divergence in the physical setup.
    This discrepancy likely arises because the simulation models the puck-table contact as a point contact, whereas real-world contact involves full surface contact, altering the friction distribution. 
    Bridging this sim-to-real gap by modeling such contact in simulation or utilizing a small amount of real-world data to learn unmodeled dynamics is a promising direction for future work. 
        
    \subsection{Edge Pushing} 
\begin{figure}[h]
        \vspace{1em}
        \centering
        \includegraphics[width=0.7\textwidth]{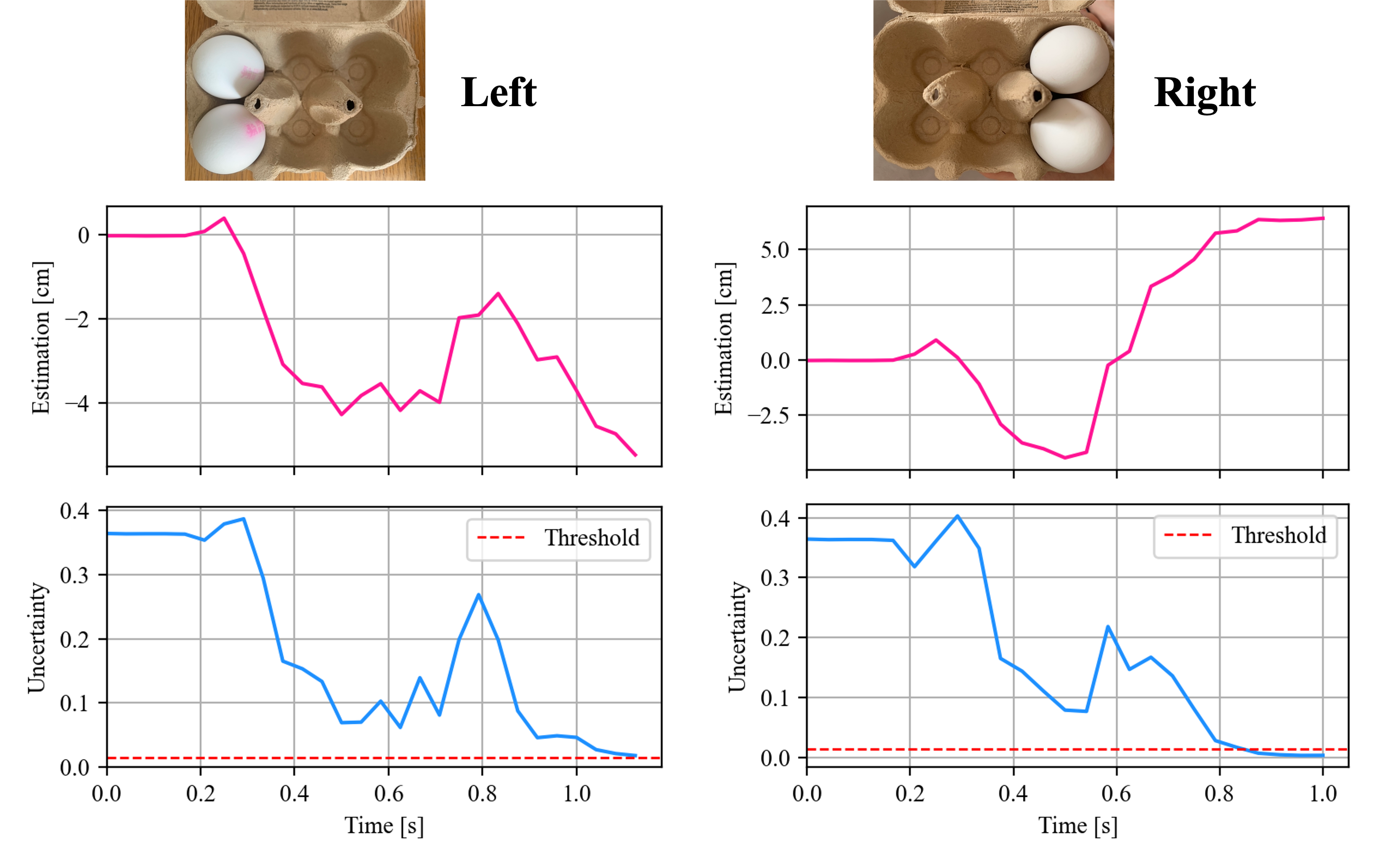}
        \caption{
        Estimates of the~$x$ component of the \textbf{center of mass} and its associated uncertainty during exploration policy execution on the physical setup for the \texttt{Edge Pushing} task.
        Uncertainty decreases below the computed threshold~$0.018$, leading to successful trials . 
        }
        \label{fig:robot_estimation_com_egg}
    \end{figure}

    \textbf{Shifted center of mass.} 
    For the \texttt{Edge Pushing} task, we tested the robot with two different center-of-mass (CoM) locations by placing the eggs on either side of the egg cartons.
    When we shifted the CoM to the left, the robot estimated its location 5.8cm off-center to the left, achieving 5/5 successful runs.
    When we shifted the CoM to the right, the robot estimated its location 5.8cm off-center to the right, achieving 3/5 successful runs.
    \cref{fig:robot_estimation_com_egg} shows that the robot’s CoM estimates converge to the correct values, with uncertainty dropping below 0.018---the computed threshold. 

\end{document}